\documentclass[sigconf]{acmart}
\usepackage{geometry}
\usepackage{graphicx}
\usepackage{float}
\usepackage{caption}
\usepackage{subfig}
\usepackage{bm}
\usepackage{multirow}

\AtBeginDocument{%
	\providecommand\BibTeX{{%
			\normalfont B\kern-0.5em{\scshape i\kern-0.25em b}\kern-0.8em\TeX}}}

\copyrightyear{2023}
\acmYear{2023}
\setcopyright{rightsretained}
\acmConference[WWW '23]{Proceedings of the ACM Web Conference 2023}{May 1--5, 2023}{Austin, TX, USA}
\acmBooktitle{Proceedings of the ACM Web Conference 2023 (WWW '23), May 1--5, 2023, Austin, TX, USA}
\acmDOI{10.1145/3543507.3583287}
\acmISBN{978-1-4503-9416-1/23/04}

\usepackage{etoolbox}
\makeatletter
\patchcmd{\maketitle}{\@copyrightpermission}{
   \begin{minipage}{0.3\columnwidth}
     \href{https://creativecommons.org/licenses/by/4.0/}{\includegraphics[width=0.90\textwidth]{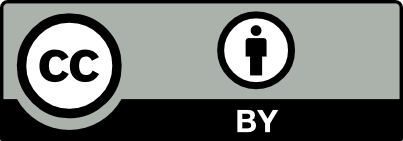}}
   \end{minipage}\hfill
   \begin{minipage}{0.7\columnwidth}
     \href{https://creativecommons.org/licenses/by/4.0/}{This work is licensed under a Creative Commons Attribution International 4.0 License.}
   \end{minipage}
  
   \vspace{5pt}
}{}{}

\makeatother

\begin{document}
	
	\title{Predicting the Silent Majority on Graphs: \\ Knowledge Transferable Graph Neural Network}
	
	\author{Wendong Bi}
	\affiliation{%
		\institution{Institute of Computing Technology, University of Chinese Academy of Sciences}
		\city{Beijing}
		\country{China}}
	\email{biwendong20g@ict.ac.cn}
	
	\author{Bingbing Xu}
	\authornote{Corresponding authors}
	\affiliation{%
		\institution{Institute of Computing Technology, Chinese Academy of Sciences}
		\city{Beijing}
		\country{China}}
	\email{xubingbing@ict.ac.cn}

	\author{Xiaoqian Sun}
	\authornotemark[1]
	\affiliation{%
		\institution{Institute of Computing Technology, Chinese Academy of Sciences}
		\city{Beijing}
		\country{China}}
	\email{sunxiaoqian@ict.ac.cn}
	
	\author{Li Xu}
	\affiliation{%
		\institution{Institute of Computing Technology, Chinese Academy of Sciences}
		\city{Beijing}
		\country{China}}
	\email{lixu@ict.ac.cn}
	
	
	\author{Huawei Shen}
	\authornotemark[1]
	\affiliation{%
		\institution{Institute of Computing Technology, Chinese Academy of Sciences}
		\city{Beijing}
		\country{China}}
	\email{shenhuawei@ict.ac.cn}
	
	\author{Xueqi Cheng}
	\authornotemark[1]
	\affiliation{%
		\institution{Institute of Computing Technology, Chinese Academy of Sciences}
		\city{Beijing}
		\country{China}}
	\email{cxq@ict.ac.cn}
	
	\renewcommand{\shortauthors}{Wendong Bi et al.}
	
	\begin{abstract}
		Graphs consisting of vocal nodes ("the vocal minority") and silent nodes ("the silent majority"), namely VS-Graph, are ubiquitous in the real world. The vocal nodes tend to have abundant features and labels. In contrast, silent nodes only have incomplete features and rare labels, e.g., the description and political tendency of politicians (vocal) are abundant while not for ordinary civilians (silent) on the twitter's social network. Predicting the silent majority remains a crucial yet challenging problem.
		However, most existing Graph Neural Networks (GNNs) assume that all nodes belong to the same domain, without considering the missing features and distribution-shift between domains, leading to poor ability to deal with VS-Graph.
		To combat the above challenges, we propose Knowledge Transferable Graph Neural Network (KTGNN), which models distribution-shifts during message passing and learns representation by transferring knowledge from vocal nodes to silent nodes.
		Specifically, we design the domain-adapted "feature completion and message passing mechanism" for node representation learning while preserving domain difference. And a knowledge transferable classifier based on KL-divergence is followed. 
		Comprehensive experiments on real-world scenarios (i.e., company financial risk assessment and political elections) demonstrate the superior performance of our method. Our source code has been open-sourced\footnote {The source code is available at \href{https://github.com/wendongbi/KT-GNN}{https://github.com/wendongbi/KT-GNN}}.

	\end{abstract}

	
	
	
	\maketitle
	
	\section{Introduction}
	Graph structured data is prevalent in the real world e.g., social networks \cite{facebook100, bonacich1987power}, financial networks \cite{zheng2021heterogeneous, cheng2019risk, yang2020financial}, and citation networks \cite{GCN, GIN, qiu2020gcc}. For many practical scenarios, the collected graph usually contains incomplete node features and unavailable labels due to reasons such as limitation of observation capacity, incompleteness of knowledge, and distribution-shift, which is referred to as the data-hungry problem for graphs. Compared with traditional ideal graphs(Fig. 1 a)), the nodes on such graphs (Fig. 1 b)) can be divided into two categories according to the degree of data-hungry: vocal nodes and silent nodes. We name such graph as the \textbf{VS-Graph}.
	
\begin{figure}[h]
		\centering
		\includegraphics[width=\linewidth]{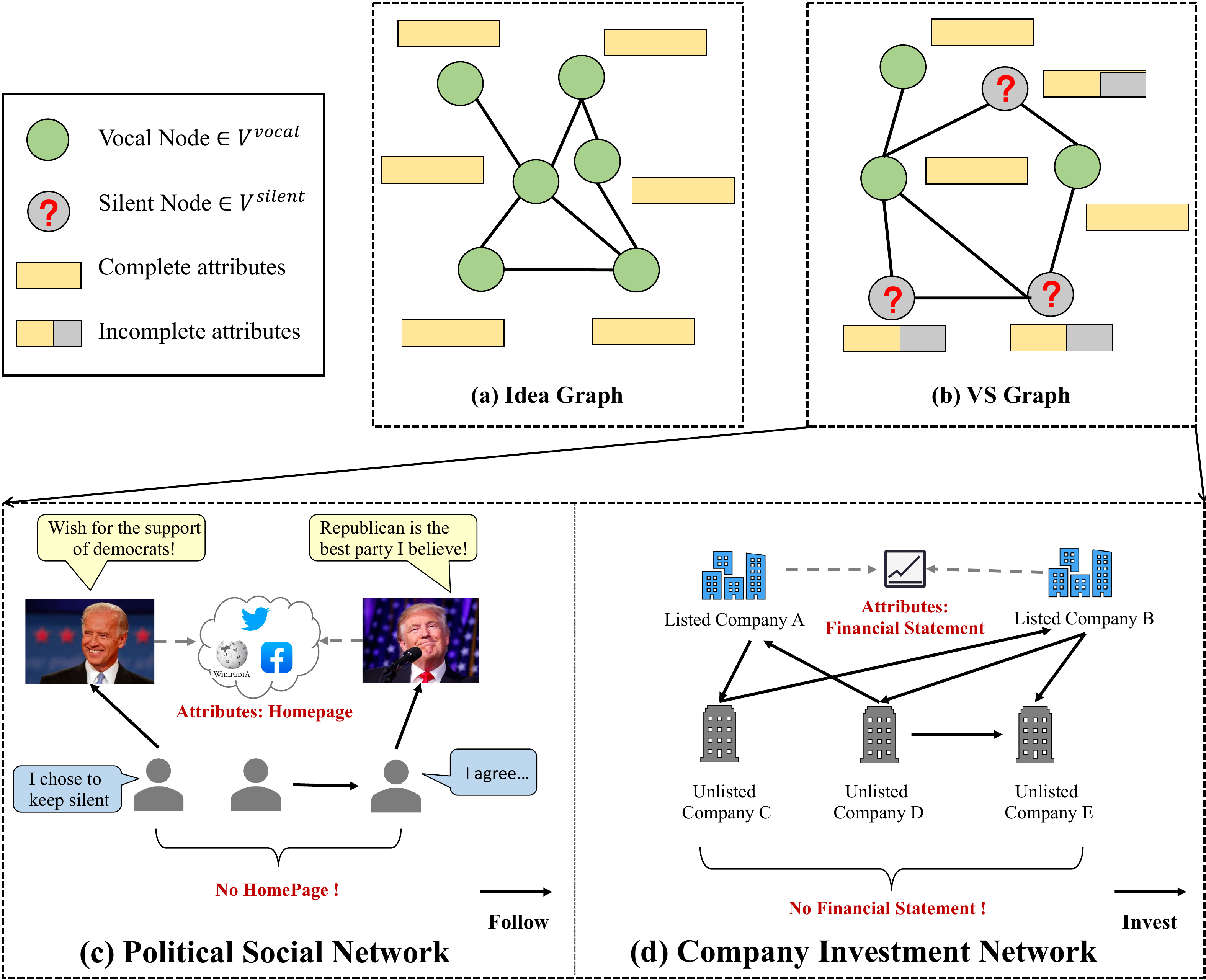}
		\caption{Examples of silent node classification, where (a) and (b) show the difference between silent node classification vs. traditional node classification. (c) and (d) show two real-world VS-Graphs.}
		\label{fig:intro}
	\end{figure}

	VS-Graph is ubiquitous and numerous graphs in the physical world. Taking two famous scenarios “political election” \cite{xiao2020timme, allen2009networks, tumasjan2011election} and “company financial risk assessment” \cite{bi2022company, zhang2021form, mai2019deep, chen2020ensemble, bougheas2015complex, feng2020every} as examples. As Fig. \ref{fig:intro} (c) illustrated, the politicians (i.e., political celebrities) in the minority and the civilians in the majority form a politician-civilian graph by social connections (e.g., following on Twitter). We can obtain detailed descriptions (attributes) and clear political tendencies (labels) for politicians (vocal nodes) while such information is unavailable for civilians (silent nodes). Meanwhile, predicting the political tendency of the majority is critical for political elections. Similar problems also exist in company financial risk assessment in Fig. \ref{fig:intro} (d), where the investment relations between listed companies in the minority and unlisted companies in the majority form the graph. Only listed companies publish their financial statements and the financial risk status of listed companies is clear. While unlisted companies are not obligated to disclose financial statements, and it is difficult to infer their risk profile based on extremely limited business information. However, the financial risk assessment of unlisted companies in the majority is of great significance for financial security \cite{liu2021pick, chaudhuri2011fuzzy, erdogan2013prediction, hauser2011predicting}. Overall, the vocal nodes (\textbf{"the vocal minority"}) have abundant features and labels. While the silent nodes (\textbf{"the silent majority"}) have rare labels and incomplete features. VS-Graph is also common in other real-world scenarios, such as celebrities and ordinary netizens in social networks. Meanwhile, predicting the silent majority on VS-Graph is important yet challenging.
	
	Recently, Graph neural networks (GNNs) have achieved state-of-the-art performance on graph-related tasks. Generally, GNNs follow the message-passing mechanism, which  aggregates neighboring nodes’ representations iteratively to update the central node representation. However, the following three problems lead to the failure of GNNs to predict the silent majority: 1) the data distribution-shift problem exists between vocal nodes and silent nodes, e.g., listed companies have more assets and cash flows and therefore show different distribution from unlisted companies in terms of attributes, which is rarely considered by the previous GNNs; 2) the feature missing problem exists and can not be solved by traditional heuristic feature completion strategies due to the above data distribution-shift; 3) the lack of labels for silent majority hinders an ideal model, and directly training models on vocal nodes and silent nodes concurrently leads to poor performance in predicting the silent majority due to the distribution-shift.
	
	To solve the aforementioned challenges, we propose \textbf{K}nowledge \textbf{T}ransferable \textbf{G}raph \textbf{N}eural \textbf{N}etwork (KTGNN), which targets at learning representations for the silent majority on graphs via transferring knowledge adaptively from vocal nodes. KTGNN takes domain transfer into consideration for the whole learning process, including feature completion, message passing, and the final classifier. Specifically, Domain-Adapted Feature Complementor (DAFC) and Domain-Adapted Message Passing (DAMP) are designed to complement features for silent nodes and conduct message passing between vocal nodes and silent nodes while modeling and preserving the distribution-shift. With the learned node representations from different domains, we propose the Domain Transferable Classifier (DTC) based on KL-divergence minimization to predict the labels of silent nodes. Different from existing transfer learning methods that aim to learn domain-invariant representations, we propose to transfer the parameters of classifiers from vocal domain to silent domain rather than forcing a single classifier to capture domain-invariant representations. Comprehensive experiments on two real-world scenarios (i.e., company financial risk assessment and political elections) show that our method gains significant improvements on silent node prediction over SOTA GNNs.
	
	The contributions of this paper are summarized as follows:
	\begin{itemize}
		\item We define a practical and widespread VS-Graph, and solve the new problem of predicting silent majority, one important and realistic AI application.
		
		\item To predict silent majority, we design a novel Knowledge Transferable Graph Neural Network (KTGNN) model, which is enhanced by the  Domain-Adapted Feature Complementor and Messaging Passing modules that preserve domain-difference as well as the KL-divergence minimization based Domain Transferable Classifier.
		
		\item Comprehensive experiments on two real-world scenarios (i.e., company financial risk assessment and political elections) show the superiority of our method, e.g., AUC achieves an improvement of nearly 6\% on company financial risk assessment compared to the current SOTA methods.
	\end{itemize}

	\section{Preliminary}
	In this section, we give the definitions of important terminologies and concepts appearing in this paper.
	\subsection{Graph Neural Network}
	Graph Neural Networks (GNNs) aim at learning representations for nodes on the graph. Given a graph $\mathbf{G}(V, E, X, Y)$ as input, where $V = \{ v_i | v_i\in V \}$ denotes the node set and $E=\{e_{i j}=(v_i, v_j) | v_i \ \text{and}\ \ v_j$ $ \text{is connected} \}$. $N=|V|$ is the number of nodes,  $X\in\mathbb{R}^{N\times D_{in}}$ is the  feature matrix and $Y\in\mathbb{R}^{N\times 1}$ is the labels of all nodes. Then GNNs update node representations by stacking multiple graph convolution layers and current mainstream GNNs follow the messaging passing mechanism, each layer of GNNs updates node representations with the following function:
	\begin{equation}
	\nonumber
	h^{(l)}_i = \mathbf{U}\left( h^{(l-1)}_i, \mathbf{M}\big(\{h^{(l-1)}_i, h_j^{(l-1)} | v_j \in \mathcal{N}(v_i) \} \big) \right)
	\end{equation}
	where $h^{(l)}_i$ is the node representation vector at $l$-th layer of GNN, $\textbf{M}$ denotes the message function of aggregating neighbor's features, and $\textbf{U}$ denotes the update functions with the neighborhood messages and central node feature as inputs. By stacking multiple layers, GNNs can aggregate information from higher-order neighbors.
	\subsection{Problem Definition: Silent Node Classification on the VS-Graph}
	In this paper, we propose the problem of silent node prediction on VS-Graph and we mainly focus on the node classification problem. First, we give the definition of VS-Graph:
	\begin{definition}[VS-Graph]
		Given a VS-Graph $\mathbf{G}^{vs}(V, E, X, Y, \Psi)$, where $V=V^{vocal}\cup V^{silent}$ is the node-set, including all vocal nodes and silent nodes. $E=\{e_{i j}=(v_i, v_j) | v_i, v_j \in  V^{vocal}\cup V^{silent}\}$ is the edge set. $X=[X^{vocal};\ X^{silent}]$ is the attribute matrix for all nodes and $Y=[Y^{vocal};\ Y^{silent}]$ is the class label of all nodes ($y_i=-1$ denotes the label of $v_i$ is unavailable). $\Psi: V\rightarrow \{vocal, silent\}$ is the population indicator that maps a node to its node population (i.e., vocal nodes, silent nodes). 
		\begin{equation}
		\label{eq:population_indicator}
		\Psi(v_i)  = \left\{
		\begin{aligned}
		& 1 \qquad \text{if $v_i$ is a silent node}\\
		& 0 \qquad \text{if $v_i$ is a vocal node}
		\end{aligned}
		\right.
		\end{equation}
	\end{definition}
	\label{sec:domain_definition}
	
	For a VS-Graph, the attributes and labels of vocal nodes (i.e., $X^{vocal}$, $Y^{vocal}$) are complete. However, $Y^{silent}$ is rare and $X^{silent}$ is incomplete. Specifically, for vocal node $v_i\in V^{vocal}$, its attribute vector $x_i=[x_i^o || x_i^u]$,  where $x_i^o\in\mathbb{R}^{D_{o}}$ is part of attributes observable for all nodes and $x_i^u\in\mathbb{R}^{D_{u}}$ is part of attributes unobservable for silent nodes while observable for vocal nodes. For silent node $v_i\in V^{silent}$, its attribute vector $x_i=[x_i^o || \mathbf{0}]$,  where $x_i^o\in\mathbb{R}^{D_{o}}$ is the observable part and $\mathbf{0}\in\mathbb{R}^{D_{u}}$ means the unobservable part which is complemented by 0 per-dimension initially. Then the problem of Silent Node Classification on VS-Graph is defined as follows:
	\begin{definition}[Silent Node Classification on the VS-Graph]
		Given a VS-Graph $\mathbf{G}^{vs}(V, E, X, Y, \Psi)$, where $|V^{vocal}| << |V^{silent}|$ and the attributes of nodes from the two populations (i.e., $X^{vocal}$ and $X^{silent}$) belong to different domains with distinct distributions $\mathcal{P}^{vocal}\neq \mathcal{P}^{silent}$. Under these settings, the target is to predict the labels ($Y^{silent}$) of silent nodes ($V^{silent}$) with the support of a small set of vocal nodes ($V^{vocal}$) with complete but out-of-distribution information.
	\end{definition}

	\section{Exploratory Analysis}
	\label{sec:data_analysis}
	\textbf{"The Silent Majority" in Real-world Graphs:} To demonstrate that the distribution-shift problem between vocal nodes and silent nodes does exist in the real-world, we choose two representative real-world scenarios (i.e., political social network, company equity network) as examples and conduct comprehensive analysis on the two real-world VS-Graphs. And the observations show that there exists a significant  distribution-shift between vocal nodes and silent nodes on their shared feature dimensions.  The detailed  information of the datasets is summarized in Sec.~\ref{sec:dataset_info}.

	\begin{figure}[h]
		\begin{minipage}[t]{0.32\linewidth}
			\centering
			\subfloat[Register capital]{\includegraphics[width=\linewidth, height=140pt]{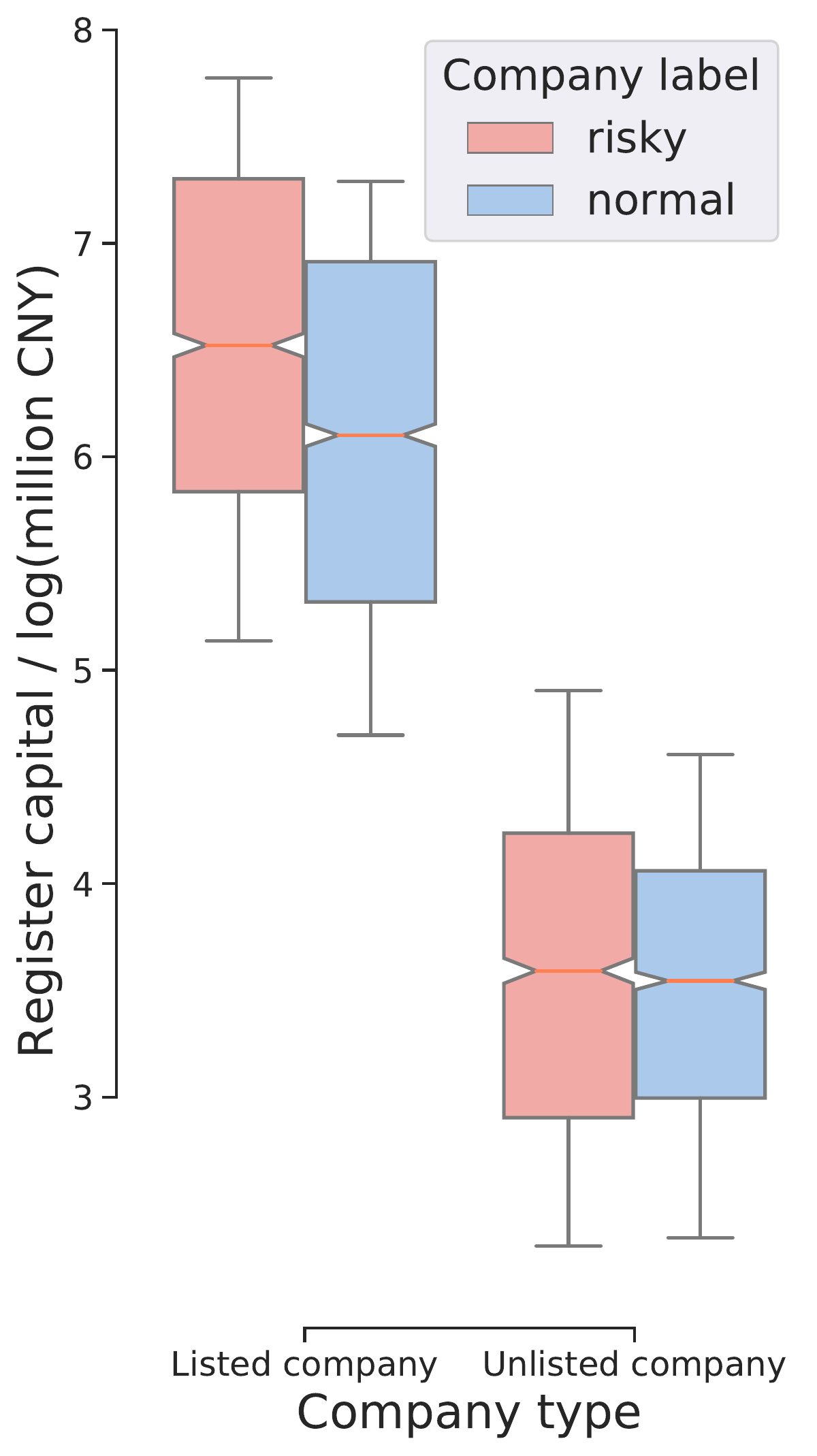}}
		\end{minipage}
		\begin{minipage}[t]{0.32\linewidth}
			\centering
			\subfloat[Actual capital]{\includegraphics[width=\linewidth, height=140pt]{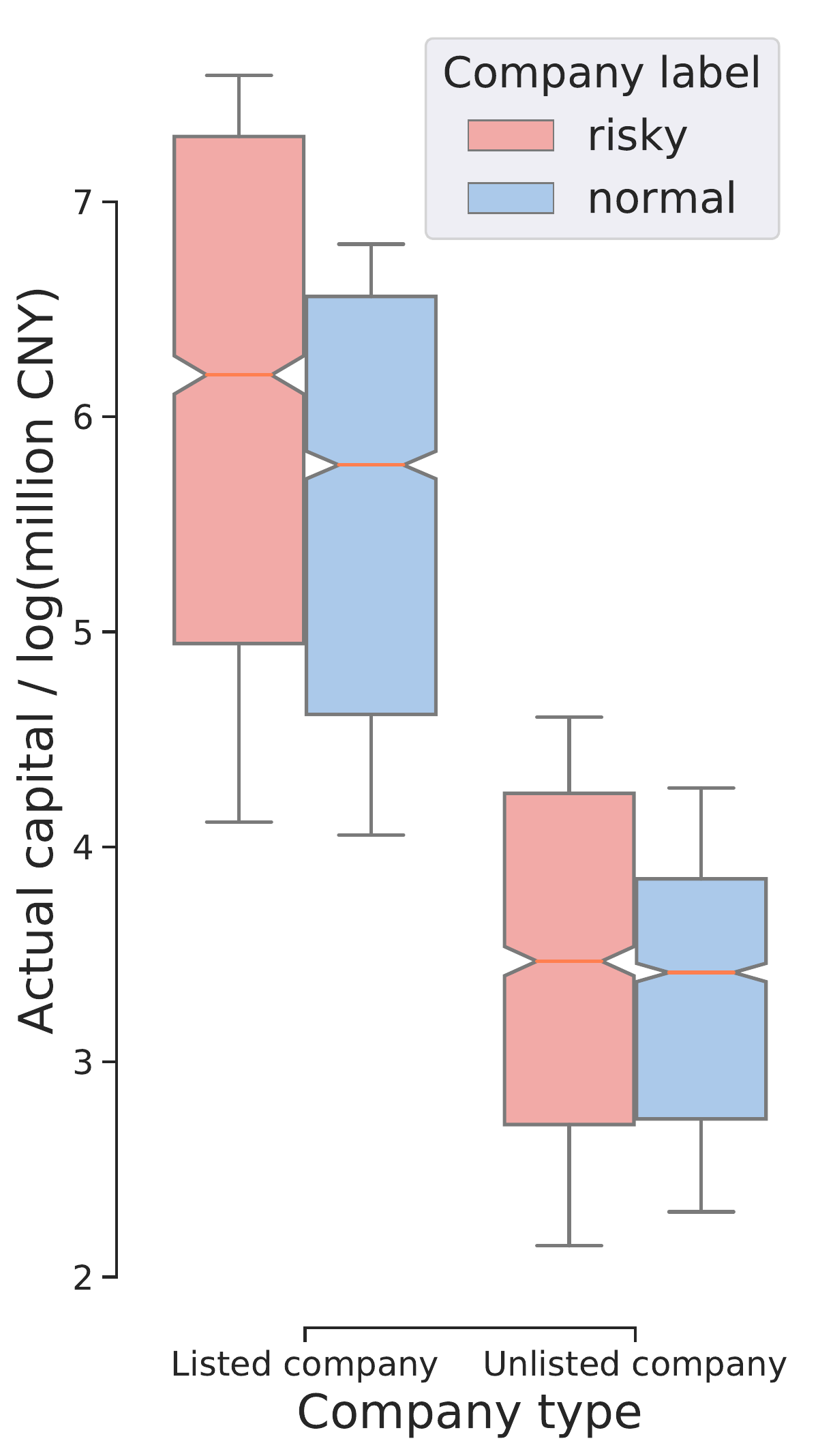}}
		\end{minipage}
		\begin{minipage}[t]{0.32\linewidth}
			\centering
			\subfloat[Staff number]{\includegraphics[width=\linewidth, height=140pt]{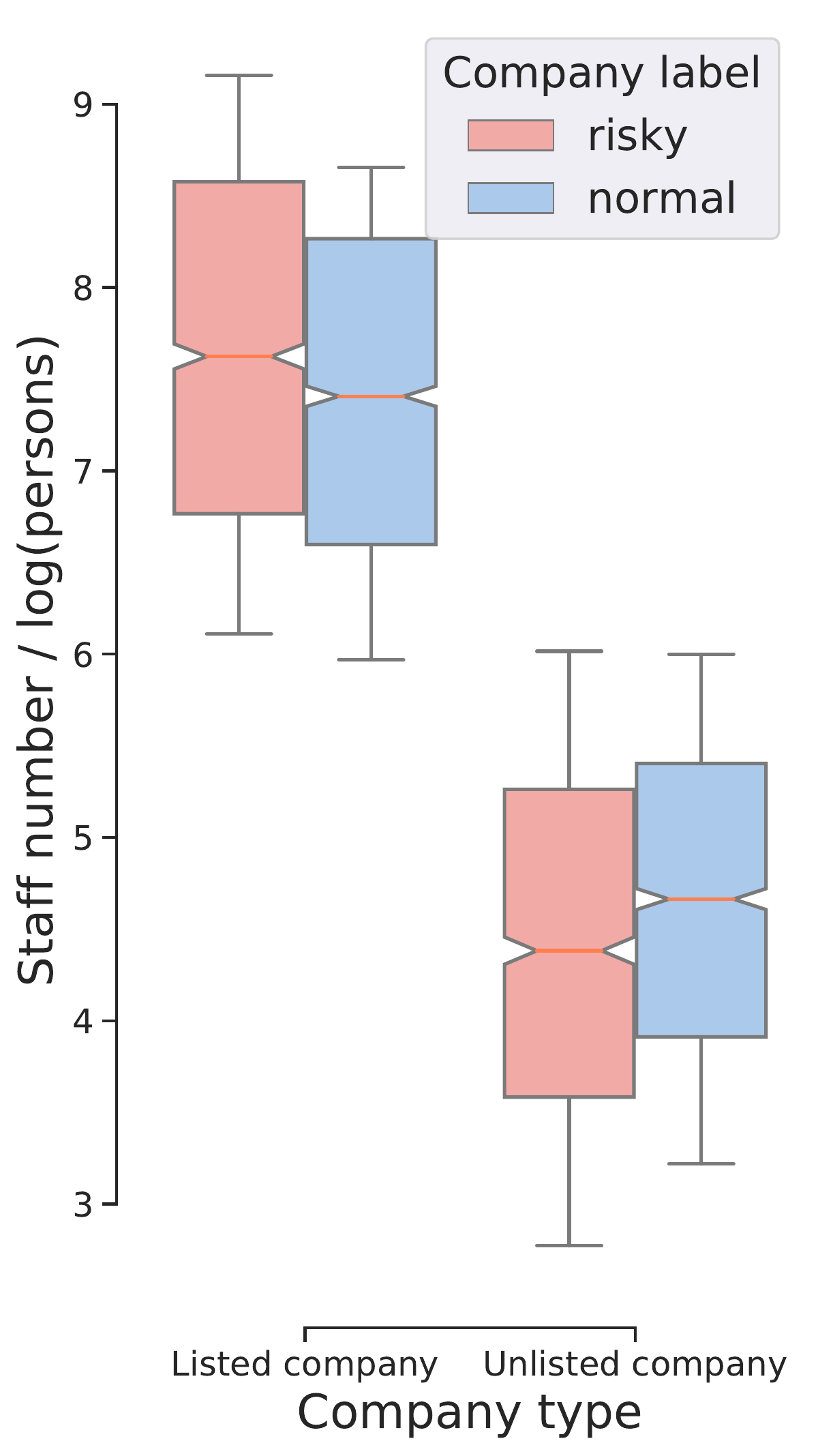}}
		\end{minipage}
		
		\caption{Box-plot of single-variable distribution for company equity graph dataset. Each box-plot actually represents the conditional distribution $P(X|Y, \mathcal{O})$, where the Y-axis is the value of attributes after the log function.}
		\label{fig:single_var_distribution}
	\end{figure}
	\subsection{Single-Variable Analysis}
	We analyze the distribution of single-variable on the real-world Company VS-Graph, including the listed company nodes and unlisted company nodes (more details in Sec. \ref{sec:dataset_info}). All node attributes of this dataset are from real-world companies and have practical physical meanings (e.g., register capital, actual capital, staff number), which are appropriate for statistical analysis of a single dimension.
	
	Instead of directly calculating the empirical distribution, we visualize the distribution with box-plots in a more intuitive manner. We select three important attributes (i.e., register capital, actual capital, staff number) and  present the box plot of all nodes distinguished by their labels (i.e., risky company or normal company) and populations (i.e., listed company or unlisted company) in the Fig.~\ref{fig:single_var_distribution}. We observe that there exists \textbf{significant distribution-shift} between listed companies (vocal nodes) and unlisted companies (silent nodes), which confirms our motivations. Considering that there are hundreds of attributes, we only present three typical attributes correlated to the company assets here due to the space limitation.

	\subsection{Multi-Variable Visualization}
	\begin{figure}[h]
		\centering
		\includegraphics[width=\linewidth]{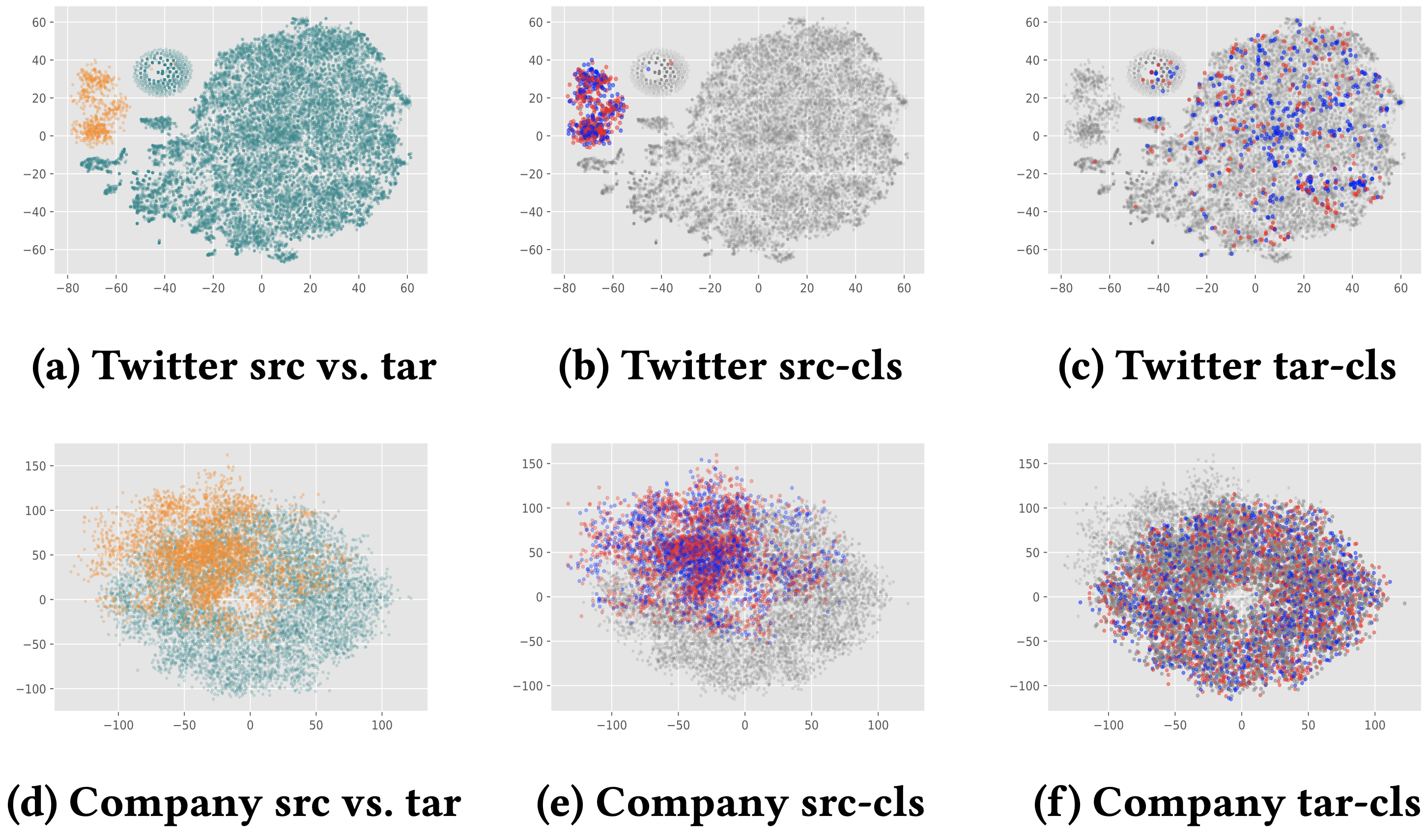}
		\caption{T-SNE visualization of the raw node features distinguished by their populations (vocal$\rightarrow$orange, silent$\rightarrow$cyan) or labels (binary classes: positive$\rightarrow$red, negative$\rightarrow$blue).  Note that we only visualize the observable attributes $X^o$ (dimensions shared by vocal and silent nodes).}
		\label{fig:tsne}
	\end{figure}
	The above single-variable analysis demonstrates that there exists significant domain difference on the certain dimension of attributes between vocal nodes and silent nodes. To further demonstrate the domain difference of all attributes, we also visualize by scatter plot on all attributes with t-SNE Algorithm \cite{van2008visualizing}. Specifically, we first project the $D_{in}$-dimensional attributes to 2-dimensional vectors and then present the scatter plot on the 2D plane.
	
	As shown in Fig.~\ref{fig:tsne}, we respectively visualize the two VS-Graph Datasets: Twitter (a political social network) and Company (a company equity network).  Specifically, Fig.~\ref{fig:tsne} (a) shows the nodes of different populations in the Twitter dataset (two colors denote vocal nodes and silent nodes); Fig.~\ref{fig:tsne} (b) shows the vocal nodes (politicians) of different classes (i.e., political parties, including democrat and republican) and Fig.~\ref{fig:tsne} (c) shows the silent nodes (civilians) of different classes. For the company dataset, Fig.~\ref{fig:tsne} (d) shows the nodes of different populations (two colors denote vocal nodes and silent nodes); Fig.~\ref{fig:tsne} (e) shows the vocal nodes (listed companies) of different classes (i.e., risky/normal company) and Fig.~\ref{fig:tsne} (c) shows the silent nodes (unlisted companies) of different classes.
	Fig.~\ref{fig:tsne} (a) and (d) demonstrate that nodes from different populations have distinct distributions, which are reflected as different clusters with distinct shapes and positions. From Fig.~\ref{fig:tsne} (b) and (e), we observe that the vocal nodes of different classes have similar distributions, which are reflected as indistinguishable clusters mixed together, and the silent nodes of different classes Fig.~\ref{fig:tsne} (c) and (f) have similar phenomena as vocal nodes. All these findings demonstrate that there exists a significant distribution-shift between vocal nodes and silent nodes on these real-world VS-Graphs.
	
	\section{Methodology}
	We propose the Knowledge Transferable Graph Neural Network (KTGNN) to learn effective representations for silent nodes on VS-Graphs.
	
	\begin{figure*}[h]
		\centering
		\includegraphics[width=0.9\linewidth]{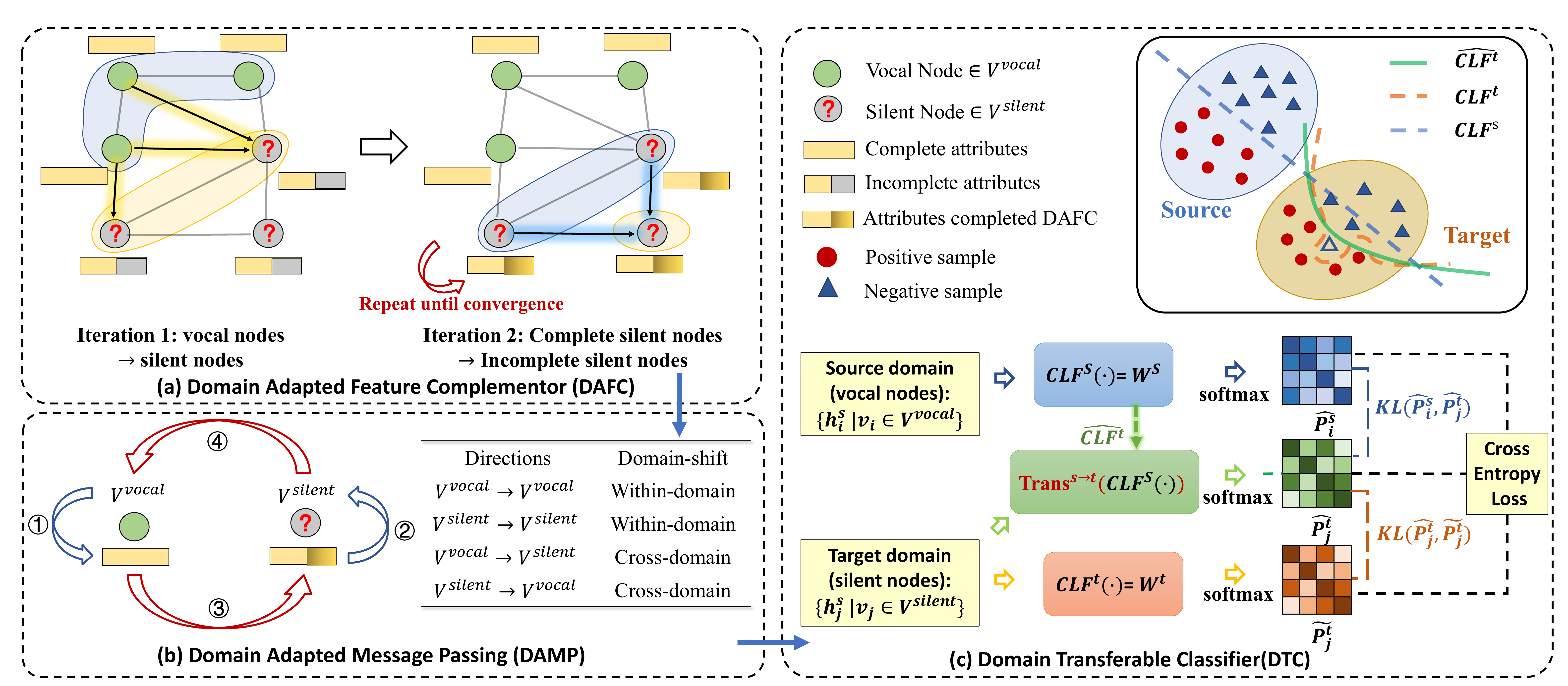}
		\caption{Architecture overview  of KTGNN, which includes three main components: Domain Adapted Feature Complementor (DAFC), Domain Adapted Message Passing (DAMP), and Domain Transferable Classifier (DTC).}
		\label{fig:model_arch}
	\end{figure*}
	\subsection{Model Overview}
	We first present an overview of our KTGNN model in Fig.~\ref{fig:model_arch}. KTGNN can be trained through an end-to-end manner on the VS-Graph and is composed of three main components: (1) Domain  Adapted Feature Complementor; (2) Domain Adapted Message Passing module; (3) Domain Transferable Classifier.
	\subsection{Domain Adapted Feature Complementor (DAFC)}
	Considering that part of the attributes are unobservable for silent nodes, we first complement the unobservable features for silent nodes with the cross-domain knowledge of vocal nodes. An intuitive idea is to complete the unobservable features of silent nodes by their vocal neighboring nodes which have complete features. However, not all silent nodes on the VS-Graph have vocal neighbors. To complete the features of all silent nodes, we propose a Domain Adapted Feature Complementor (DAFC) module which can complement the features of silent nodes by adaptively transferring knowledge from vocal nodes with several iterations.
	
	Before completing features, we first partition the nodes on the VS-Graph into two sets: $\mathcal{V}^+$ and $\mathcal{V}^-$, where nodes in $\mathcal{V}^+$ have complete features while nodes in  $\mathcal{V}^-$ have incomplete features. Initially, $\mathcal{V}^+= V^{vocal}$ and $\mathcal{V}^-=V^{silent}$. Then the DAMP module completes the features of nodes in $\mathcal{V}^-$  by transferring the knowledge of nodes in $\mathcal{V}^+$ to $\mathcal{V}^-$ iteratively. After each iteration, we add the nodes in $\mathcal{V}^-$ whose features are completed to $\mathcal{V}^+$ and remove these nodes from $\mathcal{V}^-$. And we set a hyper-parameter $K$ as the max-iteration of feature completion. By setting different $K$, the set of nodes with complete features  can cover all nodes on the Graph, and we have $V^{vocal} \subset \mathcal{V}^+ \subset V$. Fig.~\ref{fig:model_arch}-(a) present the illustration of our DAFC module. At the first iteration, features of vocal nodes are used to complete features of incomplete silent nodes that directly connect to the vocal nodes. Given incomplete silent node $v_i\in \{V^{silent}\cap \mathcal{V}^-\}$, its complemented unobservable feature $\widehat{x_i^u}$ can be calculated:
	\begin{equation}
	\label{eq:complement_iter1}
	\widehat{x_i^u} = \sum_{v_j\in\{\mathcal{N}(v_i)\cap V^{vocal}\}} \widetilde{x_j^u} \cdot f(x_i^o W^s, x_j^o W^v)
	\end{equation}
	where $\widetilde{x_j^u}$ is the calibrated variable for $x_j^u$ by eliminating the effects of domain differences:
	\begin{equation}
	\label{eq:dafc_domain_diff}
	\left\{ 
	\begin{aligned}
	\widetilde{x_j^u} &= x_j^u - \Delta\widetilde{X}^u \cdot \sigma([x_j^u || \Delta\widetilde{X}^u]\cdot W^{g})\\
	\Delta\widetilde{X}^u &= [\bar{X}^{v}_o-\bar{X}^{s}_o]\cdot W^{o\rightarrow u}
	\end{aligned}
	\right.
	\end{equation}
	where  $W^{o\rightarrow u}\in\mathbb{R}^{D^o\times D^u}$ and $W^g \in \mathbb{R}^{2D^u\times D^u}$ are learnable parametric matrix, and $\sigma$ is a $Tanh$ activation function; $\Delta\widetilde{X}^u$ is the transformed domain difference; $\bar{X}^{v}_o=\mathbb{E}_{v_i\sim V^{vocal}}(x_i^o)$ and $\bar{X}^{s}_o=\mathbb{E}_{v_i\sim V^{silent}}(x_i^o)$ are respectively the expectation of the observable features for vocal nodes and silent nodes,  and $\bar{X}^{v}_o - \bar{X}^{s}_o$ represents the domain difference  between vocal and silent nodes according to their observable attributes. In this part, we aim at learning the unobservable domain difference based on the observable domain difference $\bar{X}^{v}_o - \bar{X}^{s}_o$. Then for other incomplete silent nodes that do not directly connect to any vocal nodes, they will be complemented from the second iteration until the algorithm converges (reaching the max-iteration $K$ or all silent nodes have been complemented). After iteration 1, DAFC uses complete silent nodes to further complement features for the remaining incomplete silent nodes at each iteration. During this process, the set of complete nodes $\mathcal{V}^+$ expand layer-by-layer like breadth-first search (BFS), as shown in Eq.\ref{eq:complement_iter2}:
	\begin{equation}
	\label{eq:complement_iter2}
	\widehat{x_i^u} = \sum_{v_j\in\{\mathcal{N}(v_i)\cap \mathcal{V}^+ \}} x_j^u \cdot f(x_j^o W^v, x_i^o W^s)
	\end{equation}
	where $f(\cdot)$ is the neighbor importance factor:
	\begin{equation}
	f(x_j^o W^v, x_i^o W^s) =  \sigma\left([x_j^o W^v || x_i^o W^s] \cdot a^{vs}\right)
	\end{equation}
	where $\sigma$ is a $LeakyReLU$ activation function. Note that the node set $\{\mathcal{N}(v_i)\cap\mathcal{V}^+ \} \in V^{silent}$, because all silent nodes that have vocal neighbors have been complemented and merged into $\mathcal{V}^+$ after iteration 1 (see Eq.~\ref{eq:complement_iter1}), therefore the unobservable features $x_j^u$ in Eq.~\ref{eq:complement_iter2} do not need to be calibrated by the domain difference factor. Finally, we get the complemented features for silent nodes $\bm{\widehat{x_i} = [x_i^o || \widehat{x_i^u}]}$.
	
	To guarantee that the learned unobservable domain difference (i.e., $\Delta\widetilde{X}^u$ in Eq. \ref{eq:dafc_domain_diff}) is actually applied in the process of the domain-adapted feature completion, we add a \textbf{Distribution-Consistency Loss} $\mathcal{L}^{dist}$ when optimizing the DAFC module:
	\begin{equation}
	\label{eq:loss_dict}
	\mathcal{L}^{dist} = \left| \Delta\widetilde{X}^u - \big( \mathbb{E}_{v_i\sim V^{vocal}}(x_i^u) - \mathbb{E}_{v_i\sim V^{silent}}(\widehat{x_i^u}) \big) \right|^2
	\end{equation}
	
	\subsection{Domain Adapted Message Passing (DAMP)}
	According to the directions of edges based on the population types from source nodes to target nodes, we divide the message passing on a VS-Graph into four parts: (1) Messages from vocal nodes to silent nodes; (2) Messages from silent nodes to vocal nodes; (3) Messages from vocal nodes to vocal nodes; (4) Messages from silent nodes to silent nodes. Among the four directions of message passing, directions (1) and (2) are cross-domain (out-of-distribution) message passing, while directions (3) and (4) are within-domain (in-distribution) message passing. However, out-of-distribution messages from cross-domain neighbors should not be passed to central nodes directly, otherwise, the domain difference will become noises that degrade the model performance. 
	To solve the aforementioned problem, we design a Domain Adapted Message Passing (DAMP) module.  For messages from cross-domain neighbors, we first calculate the domain-difference scaling factor and then project the OOD features of source nodes into the domain of target nodes. e.g., for edges from vocal nodes to silent nodes, we project the features of vocal nodes to the domain of silent nodes and then pass the projected features to the target silent nodes.

	Specifically, the DAMP module considers two factors:  \textbf{the bi-directional domain difference} and  \textbf{the neighbor importance}. Given an edge $e_{i,j}=(v_i, v_j)$, the message function is given as follows:
	\begin{equation}
	\mathcal{M}_{v_i\rightarrow v_j} = \widetilde{h_i} \cdot f^{\Psi(v_j)}( \widetilde{h_i}, h_j)
	\end{equation}
	where $\widetilde{h_i}$ is the source node feature calibrated by the domain difference factor and $f^{\Psi(v_j)}(\cdot)$ is the neighbor importance function:
	
	\begin{equation}
	f^{\Psi(v_j)}(\widetilde{h_i}, h_j) =  \sigma([\widetilde{h_i}\cdot W^{\Psi(v_j)} || h_j\cdot W^{\Psi(v_j)}] )\cdot \mathbf{a}^{\Psi(v_j)}
	\end{equation}
	\begin{equation}
	\widetilde{h_i} =  \mathcal{P}_{\Psi(v_i)\rightarrow \Psi(v_j)}(h_i) = h_i + \Delta_{i,j}
	\end{equation}
	where $\sigma$ is $LeakyReLU$ and $\Delta_{i,j}$ is the distribution-shift variable:
	\begin{equation}
	\Delta_{i,j} =  \left\{\begin{aligned}
	&\qquad\qquad\qquad \mathbf{0} \qquad\qquad\qquad\qquad\quad if\ \Psi(v_i) == \Psi(v_j) \\
	&(-1)^{\Psi(v_j)}\cdot \sigma([h_i || \bar{X}^{v}-\bar{X}^{s}]\cdot \mathbf{a}^{\Psi(v_i)}) \cdot (\bar{X}^{v}-\bar{X}^{s})\quad else \\
	\end{aligned}
	\right .
	\end{equation}
	where $\sigma$ is $Tanh$; $\Psi(v_i)$ is the population indicator (see Eq.~\ref{eq:population_indicator}); $\bar{X}^{v}=\mathbb{E}_{v_i\sim V^{vocal}}(x_i)$ and $\bar{X}^{s}=\mathbb{E}_{v_i\sim V^{silent}}(\widehat{x_i})$ are respectively the expectation of the observable features for vocal nodes and silent nodes, and $\bar{X}^{v} - \bar{X}^{s}$ represents the domain difference  between vocal and silent nodes according to their complete attributes (the attributes of silent nodes have been complemented with DAFC).  With the DAMP module, we calibrate the source nodes to the domain of target nodes while message passing, which eliminates the noises caused by the OOD features. It should be noticed that the distribution-shift variable $\Delta_{i,j}$ only works for cross-domain message passing ($V^{vocal}\rightarrow V^{silent}$ or $V^{silent}\rightarrow V^{vocal}$), and $ \Delta_{i,j}=1$ for within-domain message passing ($V^{vocal}\rightarrow V^{vocal}$ or $V^{silent}\rightarrow V^{silent}$). With DAMP, we finally obtain the representations of silent nodes and vocal nodes while preserving their domain difference. Different from mainstream domain adaption methods that project the samples from source and target domains into the common space and only preserve the domain-invariance features, our DAMP method conducts message passing while preserving domain difference.

	\subsection{Domain Transferable Classifier (DTC)}
	\label{sec:dtc}
	With DAFC and DAMP, we solve the feature incompleteness problem and obtain the representations of silent nodes and vocal nodes that preserve the domain difference. Considering that the node representations come from two distinct distributions and the data-hungry problem of silent nodes (see Sec.~\ref{sec:data_analysis} and Sec.~\ref{sec:domain_definition}), we cannot directly train a good classifier for silent nodes. To solve the cross-domain problem and label scarcity problem of silent nodes,  we design a novel Domain Transferable Classifier (DTC) for the silent node classification by transferring knowledge from vocal nodes. Traditional domain adaption methods usually transfer cross-domain knowledge by constraining the learned representations to retain domain-invariance information only, under the assumption that the distribution of two domains is similar, which may not be satisfied. Rather than constraining representations, DTC targets at transferring model parameters so that the knowledge of the optimized source classifier can be transferred to the target domain.
	
	The solid rectangular box in Fig.~\ref{fig:model_arch}-(c) shows our motivation to design DTC. Specifically, the classifier trained on source domain only performs under-fitting in target domain due to domain shift (blue dotted line). Meanwhile, the classifier trained on target domain tends to overfit due to label scarcity (orange dotted line). An ideal classifier is between these two classifiers (green line). Based on this motivation (verified via experimental results in Fig.\ref{fig:loss_company}), we transfer knowledge from both source classifier and target classifier to introduce an ideal classifier by minimizing the KL divergence between them. Specifically, as shown in Fig.~\ref{fig:model_arch}-(c), DTC has three components, including the source classifier $CLF^s$, the target classifier $CLF^t$ and the cross-domain transformation module $Trans^{s\rightarrow t}$. 
 
	To predict the silent majority, the source classifier is trained with the vocal nodes and the target domain classifier is trained with the silent nodes. the cross-domain transformation module is used to transfer the original source domain classifier to a newly generated target domain classifier ($\widehat{CLF^t}=Trans^{s\rightarrow t}\left(CLF^s(\cdot) \right)$), which takes the parameters of source domain classifier as input and then generate the parameters of a new target domain classifier. Specifically, both $CLF^s=\sigma(W^sH)$ and $CLF^t=\sigma(W^tH)$ are implemented by one fully-connected layer with \textit{Sigmoid} activation. And $Trans^{s\rightarrow t}(W^s)=MLP(W^s)$ is implemented by a multi-layer perception with nonlinear transformation to give it the ability to change the shape of the discriminant boundary of the classifier.
 
	The loss function to optimize DTC and the whole model can be divided into two parts: the KL loss $\mathcal{L}^{kl}$ and the classification loss $\mathcal{L}^{clf}$. 
	
	The KL loss function $\mathcal{L}^{kl}$ is proposed to realize the knowledge transfer between the source domain and the target domain by constraining the discriminative bound of the generated classifier $\widehat{CLF^t}$ to locate between that of $CLF^{s}$  and $CLF^{t}$:
	\begin{equation}
	\label{eq:loss_adv}
	\mathcal{L}^{kl} = KL(\mathcal{P}^s, \widehat{\mathcal{P}^s}) + KL(\mathcal{P}^t, \widehat{\mathcal{P}^t})
	\end{equation}
	where $\mathcal{P}^s\in\mathbb{R}^{|V^{vocal}|\times|\mathcal{C}|}$ is the output probability of $CLF^s$, $\mathcal{P}^t\in\mathbb{R}^{|V^{silent}|\times|\mathcal{C}|}$ is the output probability of $CLF^t$, $\widehat{\mathcal{P}^s}\in\mathbb{R}^{|V^{vocal}|\times|\mathcal{C}|}$ and $\widehat{\mathcal{P}^t}\in\mathbb{R}^{|V^{silent}|\times|\mathcal{C}|}$ are the output probability of $\widehat{CLF^t}$, $|\mathcal{C}|$ is the number of classes and $|\mathcal{C}|=2$ for binary classification tasks.
	
	The classification loss $\mathcal{L}^{clf}$ is defined as: 
	\begin{equation}
	\begin{aligned}
	\mathcal{L}^{clf} = BCE(Y^{vocal}, \widehat{Y}^{vocal}_{CLF^s}) &+ BCE(Y^{silent}, \widehat{Y}^{silent}_{CLF^t})\\ &+ BCE(Y^{silent}, \widehat{Y}^{silent}_{\widehat{CLF^t}})
	\end{aligned}
	\end{equation}
	where $BCE$ is the binary cross-entropy loss function:
	\begin{equation}
	BCE(Y, \widehat{Y}, N) = \frac{1}{N} \sum_{i=1}^N Y_i \cdot \log \widehat{Y}_i + (1 - Y_i) \cdot \log (1 -\widehat {Y}_i)
	\end{equation}
	
	Combined with the Distribution-Consistency Loss $\mathcal{L}^{dict}$, our final loss function $\mathcal{L}$ is:
	\begin{equation}
	\mathcal{L} = \mathcal{L}^{clf} + \lambda\cdot \mathcal{L}^{kl} + \gamma\cdot \mathcal{L}{dist}
	\end{equation}
	where $\lambda$ is a hyper-parameter  to control the weight of $\mathcal{L}^{kl}$, and we use $\gamma=1$ in all our experiments.

	\section{Experiments}
	In this section, we compare KTGNN with other state-of-the-art methods on two real-world datasets in different scenarios.
	\subsection{Datasets}
	
	\label{sec:dataset_info}
	\begin{table}[h]
		\small
		\centering
		\setlength{\tabcolsep}{5.0pt}
		\caption{Basic information of the dataset used in this paper.}
		\label{tab:dataset_info}
		\begin{tabular}{cccccc}
			\toprule
			Dataset & $|V^{vocal}| $ & $|V^{silent}|$ & $|E|$ & $|Y^{vocal}|$ & $|Y^{silent}|$ \\
			\midrule  
			Company & 3987 &6654 & 116785  & 3987 & 1923 \\
			Twitter &581 &20230 & 915,438  & 581 & 625 \\
			\bottomrule
		\end{tabular}
	\end{table}
	
	Basic information of the real-world VS-Graphs are shown in Table \ref{tab:dataset_info}. More details about the datasets are shown in Appendix \ref{appendix:dataset}. 
	
	\textbf{Company:} Company dataset is a VS-Graph based on 10641 real-world companies in China (provided by \href{https://tianyancha.com}{TianYanCha}), and the target is to classify each company into binary classes (risky/normal). On this VS-Graph, vocal nodes denote listed companies and silent nodes denote unlisted companies. The business information and equity graphs of all companies are available, while only the financial statements of listed companies can be obtained (missing for unlisted companies). Edges of this dataset indicate investment relations between companies.
	
	\textbf{Twitter:} Following the dataset proposed by \citet{xiao2020timme}, we construct the Twitter VS-Graph based on data crawled from Twitter, and the target is  to predict the political tendency (binary classes: democrat or republican) of civilians. On this VS-Graph, vocal nodes denote famous politicians and silent nodes denote ordinal civilians. The tweet information is available for all Twitter users, while the personal descriptions from the homepage are available only for politicians (missing for civilians). Edges of this dataset represent follow relations between Twitter users.

    \subsection{Baselines and Experimental Settings}
	We implement nine representative baseline models, including $MLP$ as well as $Spectral-based$, $Spatial-based$ and $Deeper$ GNN models, which are GCN \cite{GCN}, GAT \cite{GAT}, GraphSAGE \cite{GraphSAGE}, JKNet \cite{JKNet}, APPNP \cite{APPNP}, GATv2 \cite{GATv2}, DAGNN \cite{DAGNN}, GCNII \cite{GCN2}. Besides, we choose a recent state-of-the-art GNN model (OODGAT \cite{OODGAT}) designed for the OOD node classification task.  We implement all baseline models and our KTGNN models in $Pytorch$ and $Torch-Geometric$. 
	
	For each dataset (Company and Twitter), we randomly divide the annotated silent nodes into train/valid/test sets with a fixed ratio of 60\%/20\%/20\%. And we add all annotated vocal nodes into the training set because our target is to classify the silent nodes. The detailed hyper-parameter settings of KTGNN and other baselines are presented in Appendix \ref{appendix:hyper_parameter}.
	\begin{table}[h]
		\centering
		\setlength{\tabcolsep}{1.0pt}
		\caption{Results of silent node classification. All base models except KTGNN are combined with one optimal heuristic feature completion strategy.} 
		\label{tab:main_res}
		\begin{tabular}{ccccc}
			\toprule
			Dataset &  \multicolumn{2}{c}{Twitter} & \multicolumn{2}{c}{Company}\\
			\cmidrule(lr){2-3}\cmidrule(lr){4-5}
			Evaluation Metric & F1 &  AUC  & F1  &  AUC  \\
			\midrule
			MLP & 70.85\small{$\pm$1.31} &	80.12\small{$\pm$1.38} & 57.01\small{$\pm0.42$} & 56.35\small{$\pm0.55$} \\
			GCN & 80.19\small{$\pm$0.87} & 86.88\small{$\pm$1.23} & 57.60\small{$\pm0.49$} & 57.83\small{$\pm0.93$} \\
			GAT & 82.31\small{$\pm$1.56} & 87.88\small{$\pm$1.21} & 57.98\small{$\pm$0.43} & 58.40\small{$\pm$0.43} \\
			GraphSAGE & 85.40\small{$\pm$1.03} & 92.08\small{$\pm$1.61} & 58.93\small{$\pm0.67$} & 60.63\small{$\pm0.57$} \\
			JKNet  & 83.86\small{$\pm$1.03} & 91.17\small{$\pm$1.21}  & 58.38\small{$\pm0.58$} & 61.06\small{$\pm0.53$}  \\
			APPNP & 80.60\small{$\pm$1.60} & 87.01\small{$\pm$1.08}	& 57.35\small{$\pm$0.63} & 57.99\small{$\pm$0.80} \\
			GATv2 & 82.63\small{$\pm$1.09} & 89.83\small{$\pm$1.27} & 57.78\small{$\pm$0.52} &58.93\small{$\pm$0.93} \\
			DAGNN & 84.37\small{$\pm$0.93} & 91.98\small{$\pm$1.05} & 59.62\small{$\pm$0.43} & 59.07\small{$\pm$0.39} \\
			GCNII& 83.85\small{$\pm$1.17} &	90.67\small{$\pm$1.32} & 58.21\small{$\pm0.88$} & 60.06\small{$\pm0.63$} \\
			OODGAT & 85.95\small{$\pm$2.01} &	92.67\small{$\pm$1.60} & 60.05\small{$\pm0.89$} & 61.37\small{$\pm0.92$} \\
			\midrule
			KTGNN &  $\bm{89.65}$\small{$\bm{\pm1.20}$} & $\bm{95.08}$\small{$\bm{\pm0.93}$}  & $\bm{64.96}$\small{$\bm{\pm0.63}$} & $\bm{67.11}$\small{$\bm{\pm0.52}$}  \\
			\bottomrule
		\end{tabular}
	\end{table}
	
	\subsection{Main Results}
	We focus on silent node classification on the VS-Graph in this paper and conduct comprehensive experiments on two critical real-world scenarios (i.e., political election and company financial assessment).
	
	\textbf{Results of Silent Node Classification} 
	We select two representative metrics (i.e., F1-Score and AUC) to evaluate the model performance on the silent node classification task.  Considering that baseline models cannot directly handle graphs with partially-missing features (silent nodes in VS-Graphs), we combine these baseline GNNs with some heuristic feature completion strategies (e.g., completion by zero, completion by mean of neighbors) to keep fair comparison with our methods, and we finally choose the best completion strategy for each baseline GNN (results in Table ~\ref{tab:main_res}). As shown in Table \ref{tab:main_res}, our KTGNN model gains significant improvement in the performance of silent node classification on both Twitter and Company datasets compared with other state-of-the-art GNNs. 

	\begin{table}[h]
		\centering
		\small
		\setlength{\tabcolsep}{3.0pt}
		\caption{ Results of  models with different completion strategies on Company dataset. "None" means we only use the observable  attributes (i.e., $X^o$) for both vocal and silent nodes without completion; "$0$-Completion" and  "Mean-of-Neighbors" use the zero vector and mean vector of vocal neighbors to complete the missing dimensions for silent nodes.}
		\label{tab:exp_completion_company}
		\begin{tabular}{cccc}
			\toprule
			Dataset & Completion Method & \multicolumn{2}{c}{Company}\\
			\cmidrule(lr){3-4}
			Evaluation Metric & $\backslash$ & F1 &  AUC  \\
			\midrule
			\multirow{3} *{MLP} & None & $56.06$\small{$\pm0.63$} & $55.36$\small{$\pm0.33$} \\
			~& $\bm{0}$-Completion & 56.60\small{$\pm0.67$} & 55.07\small{$\pm0.58$}	 \\
			~& Mean-of-Neighbors & 57.01\small{$\pm0.42$} & 56.35\small{$\pm0.55$}	 \\
			\hline
			\multirow{3} *{GCN} & None  & 56.61\small{$\pm0.53$} & 56.09\small{$\pm0.44$} \\
			~& $\bm{0}$-Completion & 56.41\small{$\pm0.59$} & 56.75\small{$\pm0.58$}	 \\
			~& Mean-of-Neighbors & 57.60\small{$\pm0.49$} & 57.83\small{$\pm0.93$}	 \\
			\hline
			\multirow{3} *{GraphSAGE} & None  &57.12\small{$\pm0.57$}  &57.29\small{$\pm0.57$}	\\
			~& $\bm{0}$-Completion &58.63\small{$\pm0.71$} & 59.78\small{$\pm0.65$}	 \\
			~& Mean-of-Neighbors & 58.93\small{$\pm0.67$} & 60.63\small{$\pm0.57$}	\\
			\hline
			\multirow{3} *{JKNet} & None  & 57.26\small{$\pm0.49$} & 58.44\small{$\pm0.67$}\\
			~& $\bm{0}$-Completion & 57.81\small{$\pm0.65$} & 60.95\small{$\pm0.63$}	\\
			~& Mean-of-Neighbors & 58.38\small{$\pm0.58$} & 61.06\small{$\pm0.53$} \\
			\hline
			\multirow{3} *{GCNII} & None  & 58.36\small{$\pm0.37$} & 58.85\small{$\pm0.33$}	 \\
			~& $\bm{0}$-Completion & 58.07\small{$\pm0.47$} & 59.91\small{$\pm0.56$}	 \\
			~& Mean-of-Neighbors & 58.21\small{$\pm0.88$} & 60.06\small{$\pm0.63$}	 \\
			\hline
			\multirow{3} *{OODGAT} & None  & 56.73\small{$\pm1.33$} & 57.83\small{$\pm1.20$}	 \\
			~& $\bm{0}$-Completion & 59.65\small{$\pm0.83$} & 60.93\small{$\pm0.98$} \\
			~& Mean-of-Neighbors & 60.05\small{$\pm0.89$} & 61.37\small{$\pm0.92$}	 \\
			\midrule
			KTGNN & $\backslash$ &$\bm{64.96}$\small{$\bm{\pm0.63}$} & $\bm{67.11}$\small{$\bm{\pm 0.52}$}  \\
			\bottomrule
		\end{tabular}
	\end{table}

	\textbf{Effects of Feature Completion Strategy}
	\label{sec:exp_feat_completion}
	 For baseline GNNs, we also analyze the effects of heuristic feature completion methods on the Company dataset at Table ~\ref{tab:exp_completion_twitter} (results of Twitter datasets are presented at Table ~\ref{tab:exp_completion_twitter} in Appendix).  And the results in Table ~\ref{tab:exp_completion_company} indicate that the "Mean-of-Neighbors" strategy wins in most cases. However, all these heuristic completion strategies ignore the distribution-shift between the vocal nodes and the silent nodes, and thus are far less effective than our KTGNN model.
	\begin{table}[h]
		\centering
		\setlength{\tabcolsep}{1.0pt}
		\caption{Ablation studies of KTGNN compared with its variants by removing certain components.}
		\label{tab:ablation}
		\begin{tabular}{ccccc}
			\toprule
			Dataset &  \multicolumn{2}{c}{Twitter} & \multicolumn{2}{c}{Company}\\
			\cmidrule(lr){2-3}\cmidrule(lr){4-5}
			Evaluation Metric & F1 &  AUC  & F1  &  AUC  \\
			\midrule
			$\text{KTGNN}_{\backslash\text{DAFC}}$ & 85.57\small{$\pm$1.38} &	92.12\small{$\pm$0.87} & 60.21\small{$\pm0.39$} & 61.95\small{$\pm0.45$} \\
			$\text{KTGNN}_{\backslash\text{DAMP}}$ & 86.85\small{$\pm$1.01} &	93.03\small{$\pm$0.98} & 62.71\small{$\pm0.44$} & 64.27\small{$\pm0.61$} \\
			$\text{KTGNN}_{\backslash\text{DTC}}$ & 88.80\small{$\pm$0.93} &	94.13\small{$\pm$1.08} & 63.05\small{$\pm0.52$} & 64.35\small{$\pm0.55$} \\
			$\text{KTGNN}_{\backslash\mathcal{L}^{dist}}$ & 87.80\small{$\pm$0.83} &	93.85\small{$\pm$1.38} & 63.56\small{$\pm0.68$} & 65.38\small{$\pm0.65$} \\
			$\text{KTGNN}_{\backslash\mathcal{L}^{kl}}$ & 89.20\small{$\pm$1.20} &	94.06\small{$\pm$1.03} & 62.71\small{$\pm0.63$} & 65.02\small{$\pm0.55$} \\
			\midrule
			KTGNN &  $\bm{89.65}$\small{$\bm{\pm1.20}$} & $\bm{95.08}$\small{$\bm{\pm0.93}$}  & $\bm{64.96}$\small{$\bm{\pm0.63}$} & $\bm{67.11}$\small{$\bm{\pm0.52}$}  \\
			\bottomrule
		\end{tabular}
	\end{table}
        \begin{figure}[h]
		\begin{minipage}[t]{0.325\linewidth}
			\centering
			\subfloat[F1-score]{\includegraphics[width=\linewidth]{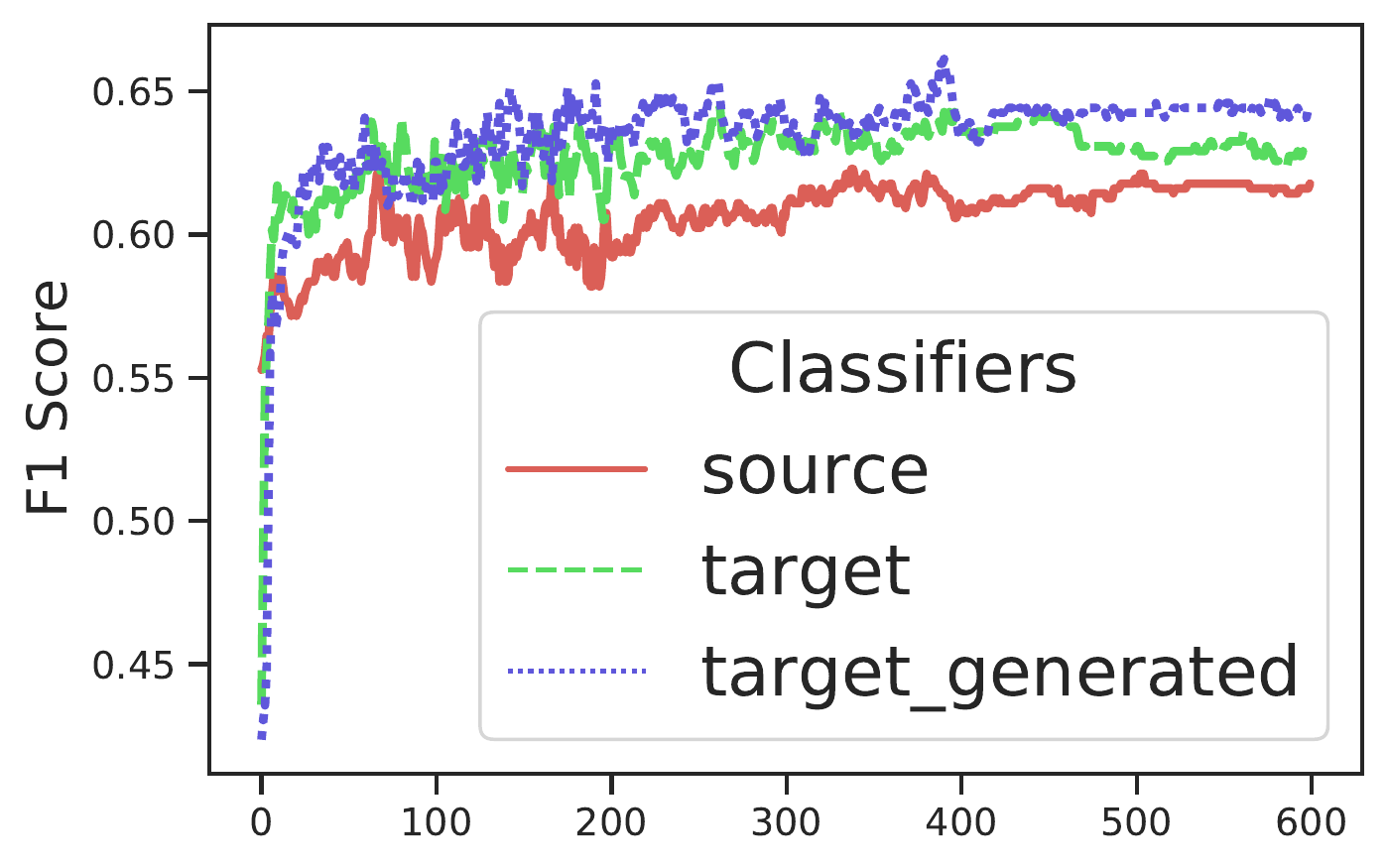}}
		\end{minipage}
		\begin{minipage}[t]{0.325\linewidth}
			\centering
			\subfloat[BCE Loss]{\includegraphics[width=\linewidth]{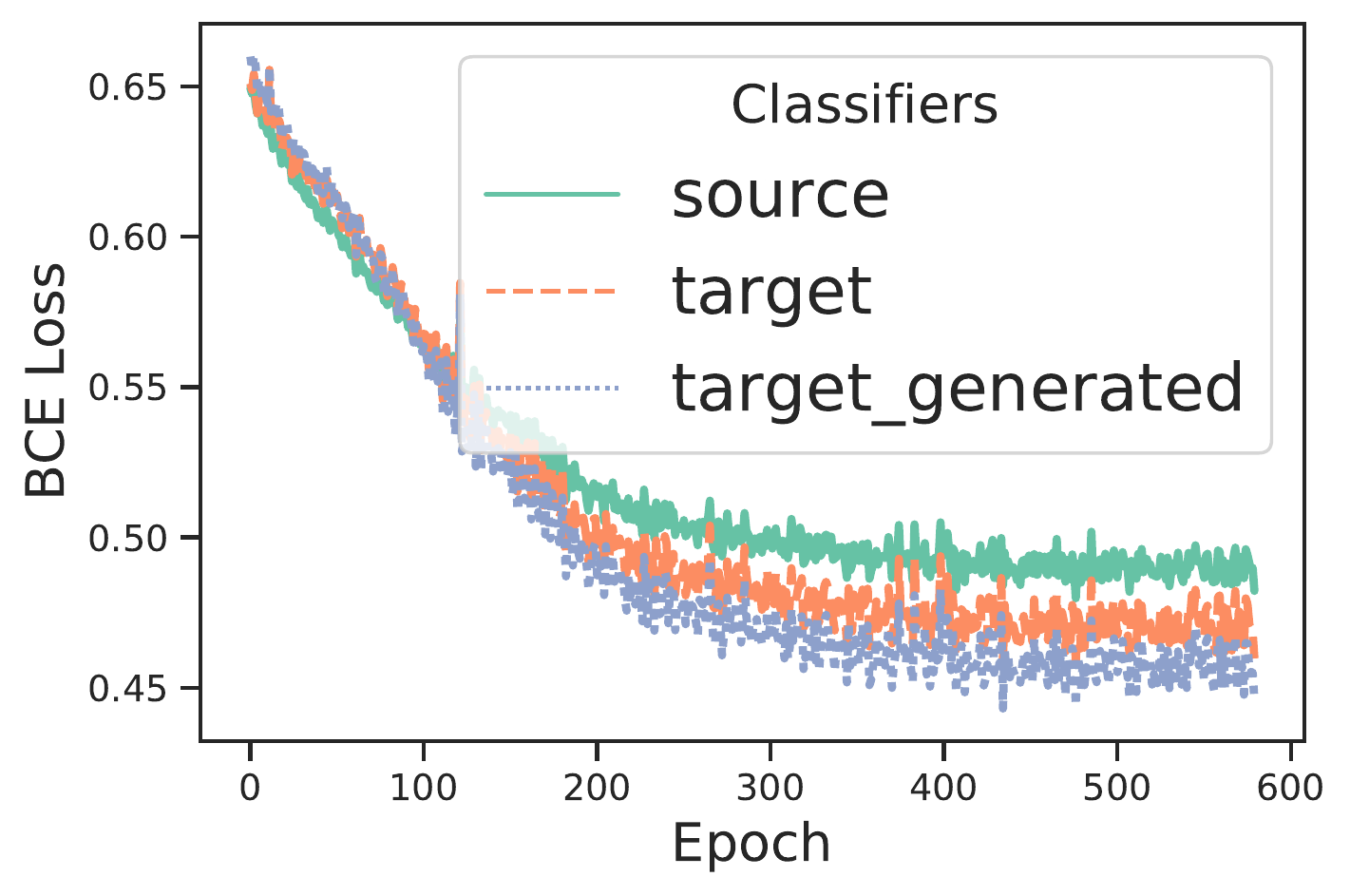}}
		\end{minipage}
		\begin{minipage}[t]{0.325\linewidth}
			\centering
			\subfloat[KL Loss]{\includegraphics[width=\linewidth]{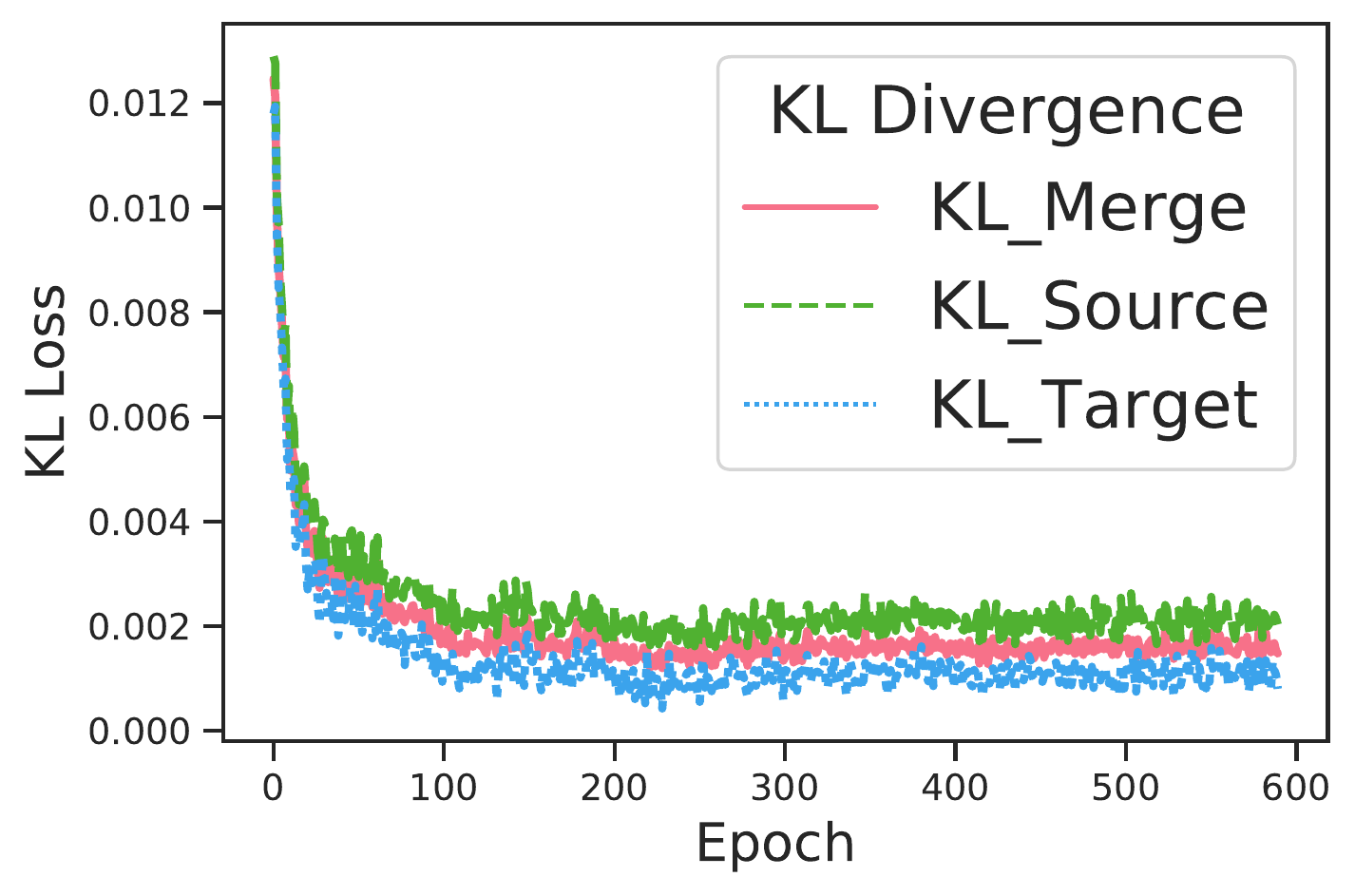}}
		\end{minipage}
		
		\caption{F1 Score and Loss Curve of KTGNN on Company dataset (results of Twitter dataset are shown in appendix).}
		\label{fig:loss_company}
	\end{figure}
	\subsection{Ablation Study}\quad
	\label{sec:ablation}
	\textbf{Effects of Each Module: } 
	To validate the effects of each component of KTGNN, we design some variants of KTGNN by removing one of modules (i.e., DAFC, DAMP, DTC, $\mathcal{L}^{dist}$, $\mathcal{L}^{kl}$) and the results are shown in the Table \ref{tab:ablation}. We observe that the performance of all KTGNN variants deteriorates to some degree. And our full KTGNN module gains the best performance, which demonstrates the important effects of each component in KTGNN.
	
	\textbf{Effects of KL Loss: }
	We also validate the role of KL loss $\mathcal{L}^{kl}$ (see Eq. \ref{eq:loss_adv}) in our method. As shown in the Fig. \ref{fig:loss_company},  the final target classifier generated from the source classifier gains the highest scores and lowest loss among the three sub-classifiers in the DTC module (see Sec.~\ref{sec:dtc}). And the result of KL Loss (components of $\mathcal{L}^{kl}$) indicates that the discriminant bound of the generated target classifier is located between that of the source classifier and target classifier, which further confirms our motivations.

\textbf{Effects of Cross-Domain Knowledge Transfer:} We further validate our approach with borderline cases (i.e., topological traits that may worsen the correct knowledge transfer from vocal nodes to silent nodes). We validate this by cutting part of such edges. The results in Table \ref{tab:exp_boarderline} indicate that our approach is robust over these borderline cases.
        \begin{figure}[h]
		\begin{minipage}[t]{0.49\linewidth}
			\centering
			\subfloat[Company-$K$]{\includegraphics[width=\linewidth, height=70pt]{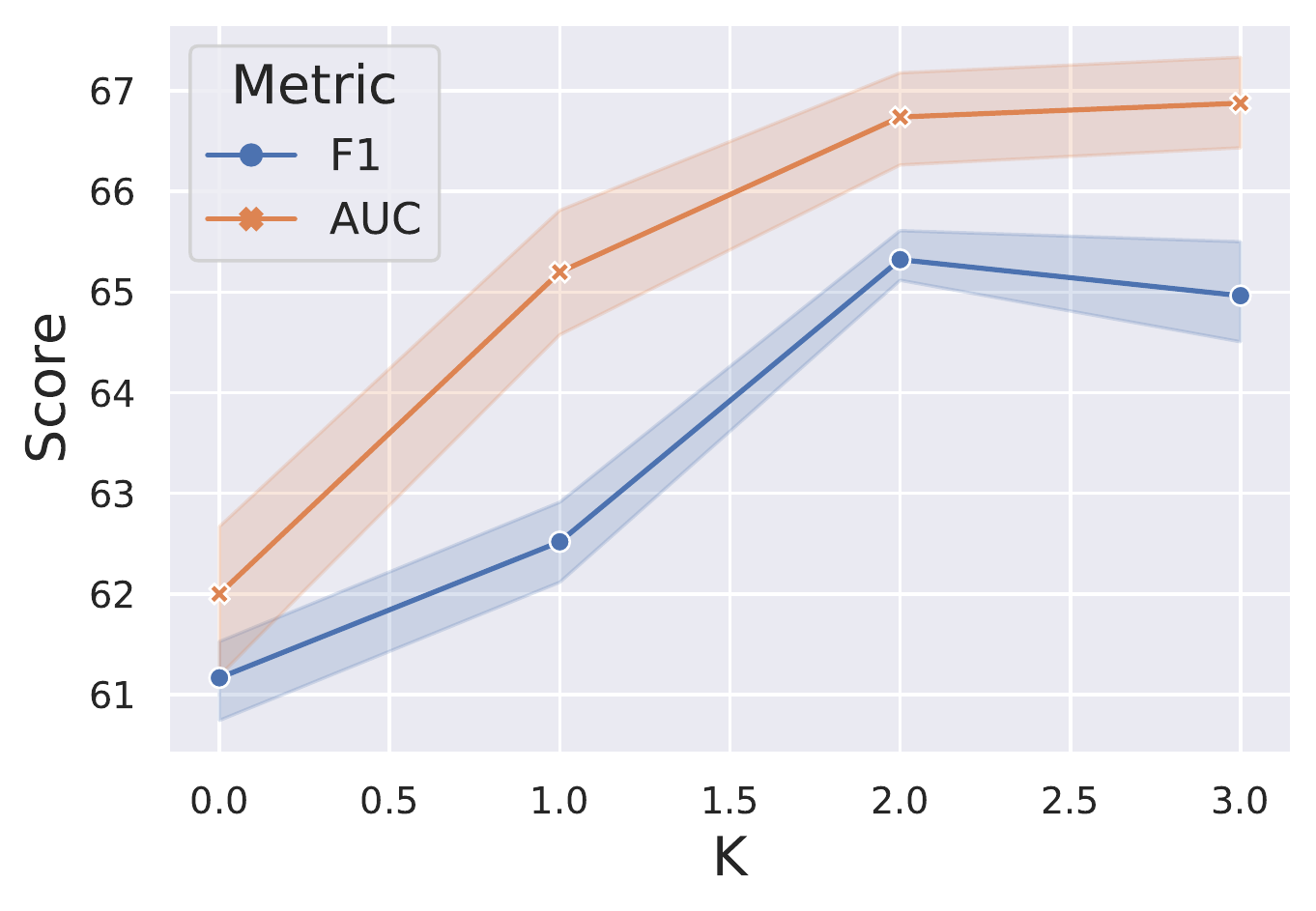}}
		\end{minipage}
		\begin{minipage}[t]{0.49\linewidth}
			\centering
			\subfloat[Company-$\lambda$]{\includegraphics[width=\linewidth, height=70pt]{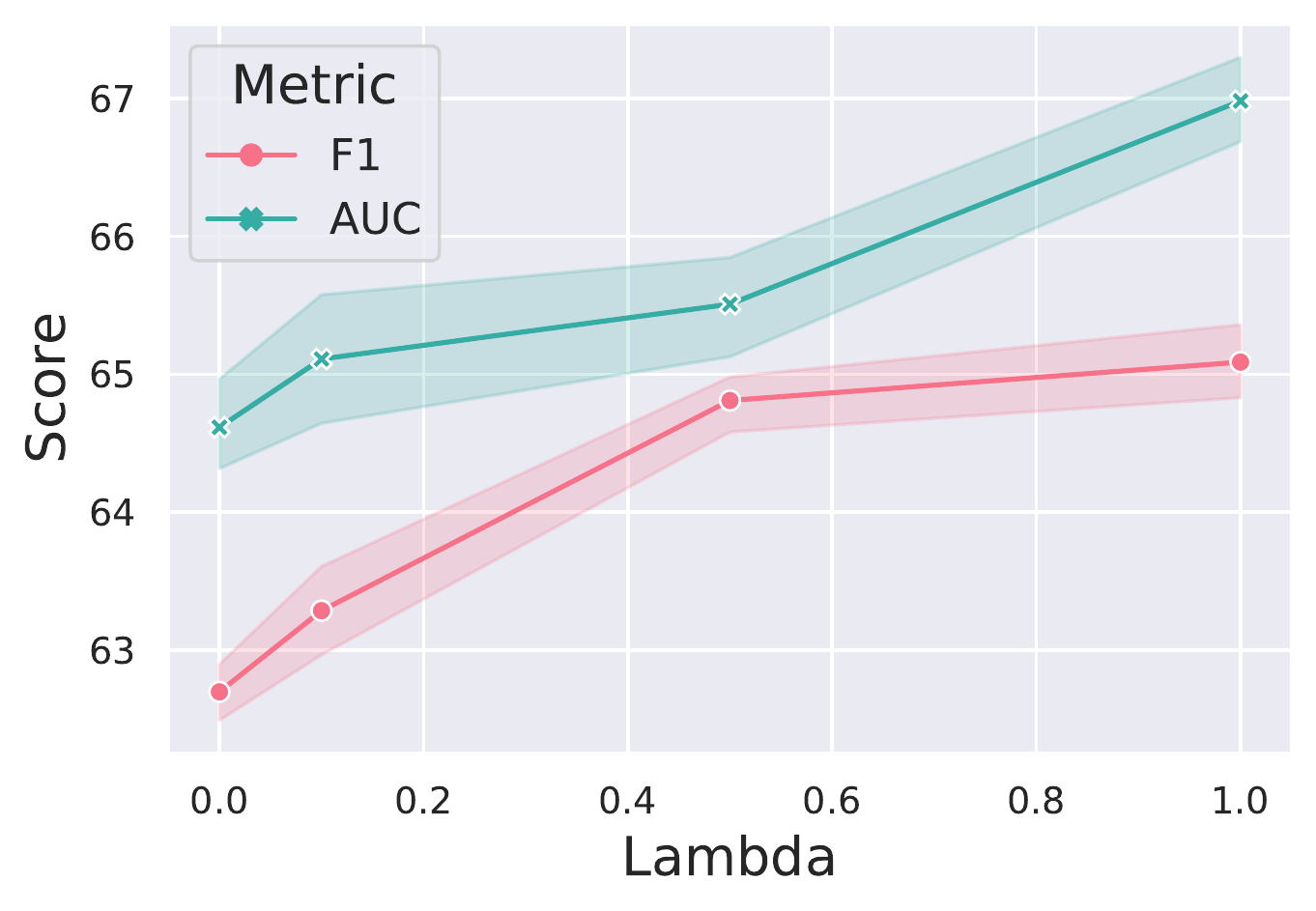}}
		\end{minipage}
		
		\caption{Hyper-parameter analysis of $K$ and $\lambda$ on Company dataset (see results on Twitter dataset in Fig. ~\ref{fig:hyper_twitter} of Appendix).}
		\label{fig:hyper-company}
	\end{figure}
        \begin{table}[h]
		\centering
		\setlength{\tabcolsep}{1.0pt}
		\caption{Boarder-line cases analysis on the effect of cross-domain edges by removing part (i.e., 0\%, 30\%, 50\%, and 70\%) of such edges randomly from the original VS-Graph.}
		\label{tab:exp_boarderline}
		\begin{tabular}{ccccc}
			\toprule
			Removing \#\% of  &  \multicolumn{2}{c}{Twitter} & \multicolumn{2}{c}{Company}\\
			\cmidrule(lr){2-3}\cmidrule(lr){4-5}
			 cross-domain edges & F1 &  AUC  & F1  &  AUC  \\
			\midrule
			 $70\%$ & 86.40\small{$\pm$1.18} &	92.33\small{$\pm$0.96} & 61.73\small{$\pm0.70$} & 63.34\small{$\pm0.65$} \\
			 $50\%$ & 88.00\small{$\pm$1.03} &	93.53\small{$\pm$1.03} & 62.91\small{$\pm0.59$} & 64.89\small{$\pm0.58$} \\
			$30\%$  & 88.80\small{$\pm$1.29} &	94.56\small{$\pm$0.97} & 63.70\small{$\pm0.73$} & 66.05\small{$\pm0.61$} \\
			\midrule
			$0\%$ &  $\bm{89.65}$\small{$\bm{\pm1.20}$} & $\bm{95.08}$\small{$\bm{\pm0.93}$}  & $\bm{64.96}$\small{$\bm{\pm0.63}$} & $\bm{67.11}$\small{$\bm{\pm0.52}$}  \\
			\bottomrule
		\end{tabular}
	\end{table}
 
	\subsection{HyperParameter Sensitivity Analysis}
	\label{sec:hyper_analysis}
	
	We validate the sensitivity of main hyper-parameters of  KTGNN: \textbf{$K$ (max-iteration of DAFC )}  and \textbf{ $\lambda$ (weight of $\mathcal{L}^{kl}$)}. As shown in Fig.~\ref{fig:hyper-company}, we conduct experiments on the Company dataset with different $K$ and $\lambda$. The results show that KTGNN gains the best performance when $K=2$, which is decided by the graph property that 2-hop neighbors of vocal nodes cover most silent nodes, and gains the best performance when $lambda=1.0$, which also indicates the effectiveness of the KL loss.
	\subsection{Representation Visualization}
	\label{sec:exp_representation}
	
\begin{figure}[h]
		\centering
		\includegraphics[width=\linewidth]{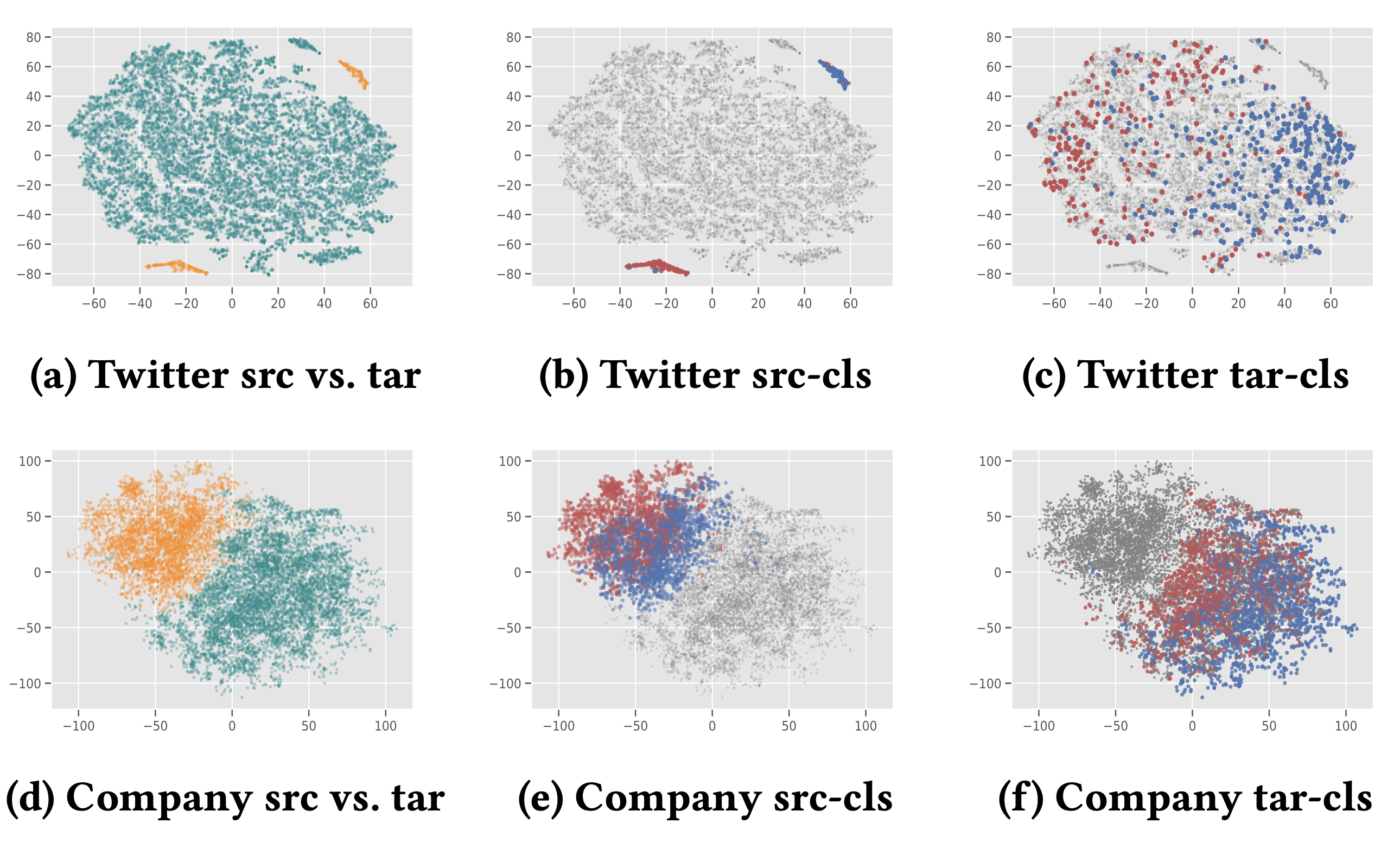}
		\caption{T-SNE visualization of the node completed representations learned by KTGNN, distinguished by their populations (vocal$\rightarrow$orange, silent$\rightarrow$cyan) or node labels (binary classes: positive$\rightarrow$red, negative$\rightarrow$blue).}
		\label{fig:tsne_representation}
	\end{figure}

	
	We visualize the learned representations of KTGNN in Fig. \ref{fig:tsne_representation}. Sub-figures (a) and (d) show that our KTGNN remains the domain difference between vocal nodes and silent nodes during message-passing. And Sub-figures (b)~(c) and (d)~(f) show that the representations learned by KTGNN have better distinguishability between different classes for both vocal nodes and silent nodes, compared with the T-SNE visualization of raw features in Fig.~\ref{fig:tsne} (The representation visualization of baseline GNNs are shown in the appendix).
	\section{Related Work}
	
	
	
	
	Graph Neural Networks (GNNs), as powerful tools for modeling graph data, have been widely proposed and can be generally categorized into \textit{spectral-based methods} \cite{ChebNet, defferrard2016convolutional, gwnn, chen2017stochastic, huang2018adaptive} and \textit{spatial-based methods} \cite{GAT, xu2020graph, Sun2011trading,xu2020label, wsdm2022guo, Guo2022EvolutionaryPL, Zhang2022EfficientlyLM, lin2020initialization, lin2014large, du2021gbk, Sun2014predict}. Existing mainstream spatial-based GNNs \cite{ xu2021towards, du2021tabularnet, du2018galaxy, bi2022make, zhang2018end, yue2022htgn, sun2017detecting, shi2019detect, GAE} follow the message-passing framework, where node representations are updated based on features aggregated from neighbors. Recently, deeper GNNs \cite{GCN2, bi2022mm, li2021training_gcn_1000layers} with larger receptive fields, have been proposed and gained better performance.  However, most of the existing GNNs are built on the I.I.D. hypothesis, i.e., training and testing  data are independent and identically distributed. And the I.I.D. hypothesis is hard to be satisfied in many real-world scenarios where the model performance degrades significantly when there exist distribution-shifts among the nodes on the graph.
	
	O.O.D nodes commonly exist in real-world graphs \cite{mao2021source, fan2021generalizing, mao2021neuron, fu2021neuron, yang2020factorizable, ma2019disentangled, AKGR, SymCLKGE}, which bring obstacles to current message-passing based GNN models. Recently, GNNs for graph-level O.O.D prediction \cite{li2022ood, sadeghi2021distributionally, gui2022good,wang2022deconfounding, wu2022discovering, huang2019dot, wu2023energybased, ligraphde} have been proposed. However, GNNs for semi-supervised node classification with O.O.D nodes have been rarely studied. Some existing studies \cite{OODGAT, wang2021confident, huangend, wu2022towards, fan2022debiased, yang2022learning} combine GNN with the O.O.D detection task as an auxiliary to enhance the performance of GNNs on graphs with O.O.D nodes. Such kind of method usually views O.O.D nodes as noises, which aim at detecting potential O.O.D nodes first and then alleviating their effects. However, O.O.D nodes are not noises in all scenarios (e.g., VS-Graph introduced in this paper) and knowledge from other domains can be helpful to alleviate the data-hungry problem of the target domain. Different from existing O.O.D detection methods, we propose to improve the model performance on the target domain by injecting out-of-distribution knowledge in this paper.
	
	\section{Conclusion}
	In this paper, we propose a novel and widespread problem: silent node classification on the VS-Graph ("predicting the silent majority on graphs"), where the silent nodes suffer from serious data-hungry problems (feature-missing and label scarcity). Correspondingly, we design a novel KTGNN model for silent node classification by transferring useful knowledge from vocal nodes to silent nodes adaptively. Comprehensive experiments on the two representative real-world VS-Graphs demonstrate the superiority of our approach. 

    \begin{acks}
		This work was supported by the National Natural Science Foundation of China (Grant No.U21B2046, No.62202448, No.61902380, No.61802370,  and No.U1911401), the Beijing Nova Program (No. Z201100006820061) and China Postdoctoral Science Foundation (2022M713208).
	\end{acks}

	
	
	

    \newpage
	\bibliographystyle{ACM-Reference-Format}
	\bibliography{sample-base}
 
	\newpage
	\appendix
	\section{Dataset Supplementary Information}
	\label{appendix:dataset}
    \subsection{Company Dataset } 
	The Company dataset is a real-world company equity graph based on companies in China\footnote{Company dataset is provided by TianYanCha:\href{https://tianyancha.com}{https://tianyancha.com}}, and the target is to predict financial risk status of companies. And the node features come from two parts: business information and financial statement. Only listed companies provide financial statements while we cannot obtain that of unlisted companies because only listed companies have disclosure obligations of financial statements. Besides the feature-missing problem, there also exists a distinct distribution-shift between listed companies and unlisted companies, which has been discussed in Sec.~\ref{sec:data_analysis}, and this leads to significant domain differences. 
	We select 33 dimensions from business information and 78 dimensions from the financial statement, and we further perform feature normalization on them. Specifically, for both business information and financial statement features, we remove the column whose missing rate is larger than 50\%, and then we complete the missing attributes by zero. Besides, we use the quantile method to truncate the outlier features with the 10\% quantile and the 90\% quantile for each column. And we process the categorical or discrete attributes into one-hot vectors, and then we use binning methods to divide continuous numerical attributes into 50 bins and use the index of the bin as their feature. Then we conduct z-score normalization on each column of the attributes.  Finally, we retain 111-dimensional  features (33 dimensions as observable features and 78 dimensions as unobservable features).
 
\subsection{Twitter Dataset }
	Twitter social networks \cite{rossi2020temporal, drakopoulos2021graph, larsson2012studying} are widely used for the political election prediction problem \cite{fraia2014use, bermingham2011using, shin2017political, guarino2020characterizing}. Following \citet{xiao2020timme}, we construct the VS-Graph based on their proposed dataset \footnote{The raw Twitter dataset is available at: \\ \href{https://github.com/PatriciaXiao/TIMME/tree/master/data}{https://github.com/PatriciaXiao/TIMME/tree/master/data}}. The Twitter datasets is a political social network dataset crawled from Twitter and the target is to predict the political tendency of civilians on this network.  For politicians (vocal nodes), we use both their tweet embedding and their account description/status fields as their attributes. For civilians (silent nodes), we only use their tweets embedding as their attributes without using their account description/status fields. And we connect politicians and civilians by their "follow" relation on Twitter and construct the Twitter VS-Graph. All node features, including tweets and personal descriptions, are text data, therefore  these words are embedded with the GloVe algorithm. Finally, we have 1836 dimensional features (300 dimensions as observable features and 1536 dimensions as unobservable features).

\subsection{Clarification on the definition of VS-Graph} 
Based on our problem setting, whether the node belongs to vocal/silent groups is determined by its inherent semantic meanings (e.g., politician/civilian, listed/unlisted company). Besides, our problem setting is also common in other real scenarios (e.g., star/non-star researchers on co-author networks by their citations, active/inactive users in recommendation scenarios by their daily activeness). 
 \begin{figure}[h]
		\begin{minipage}[t]{0.325\linewidth}
			\centering
			\subfloat[F1-score]{\includegraphics[width=\linewidth]{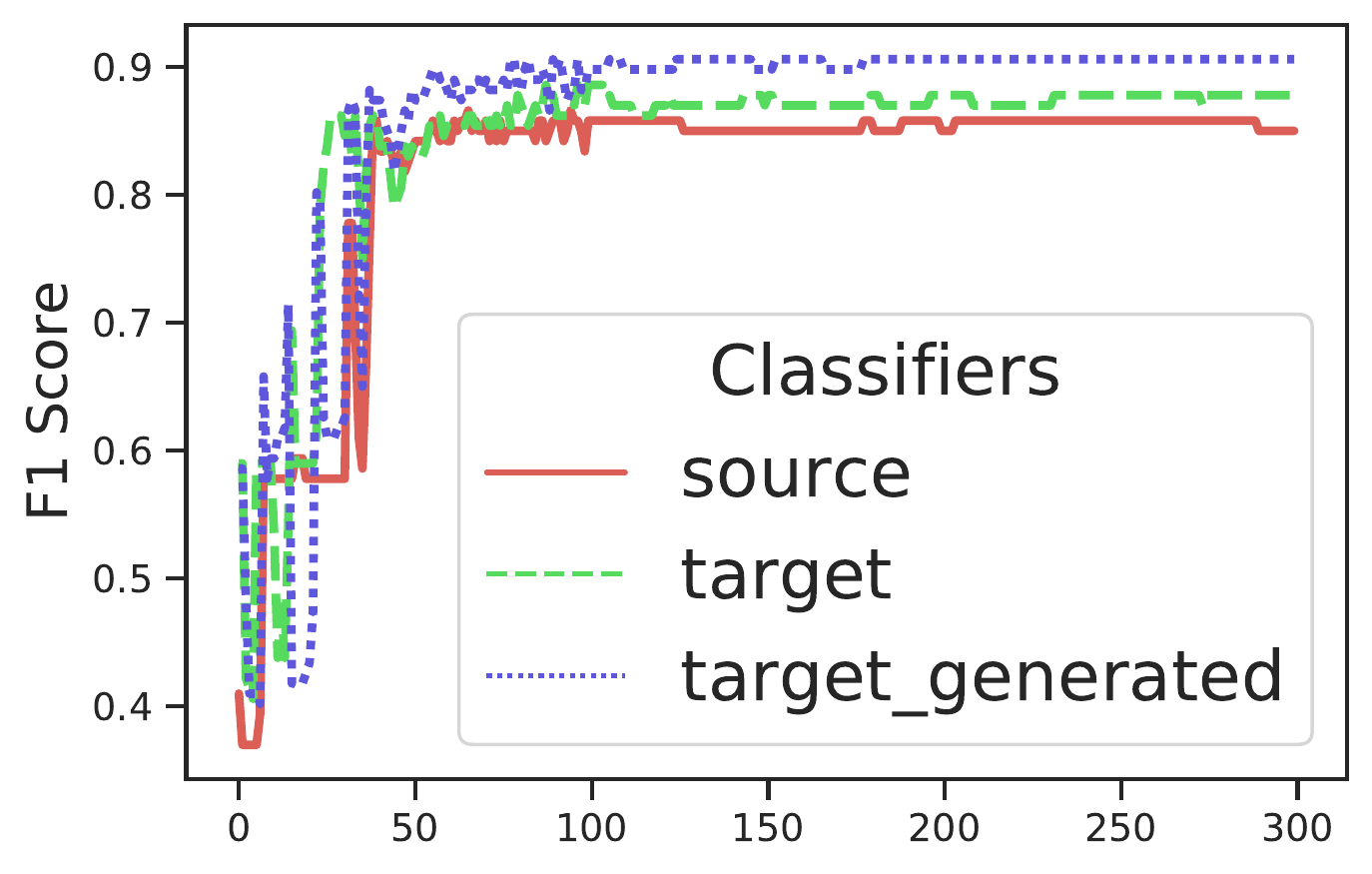}}
		\end{minipage}
		\begin{minipage}[t]{0.325\linewidth}
			\centering
			\subfloat[BCE Loss]{\includegraphics[width=\linewidth]{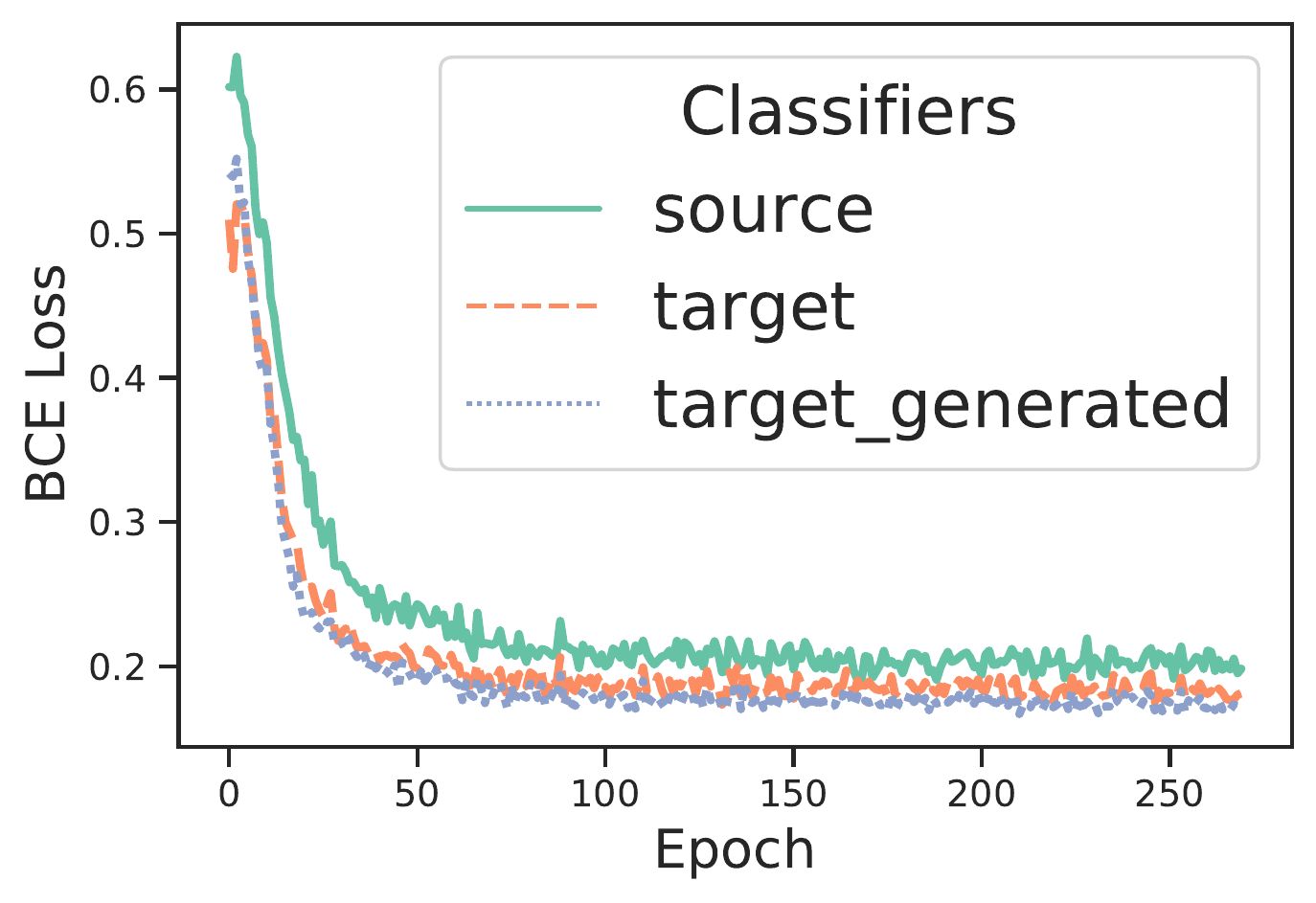}}
		\end{minipage}
		\begin{minipage}[t]{0.325\linewidth}
			\centering
			\subfloat[KL Loss]{\includegraphics[width=\linewidth]{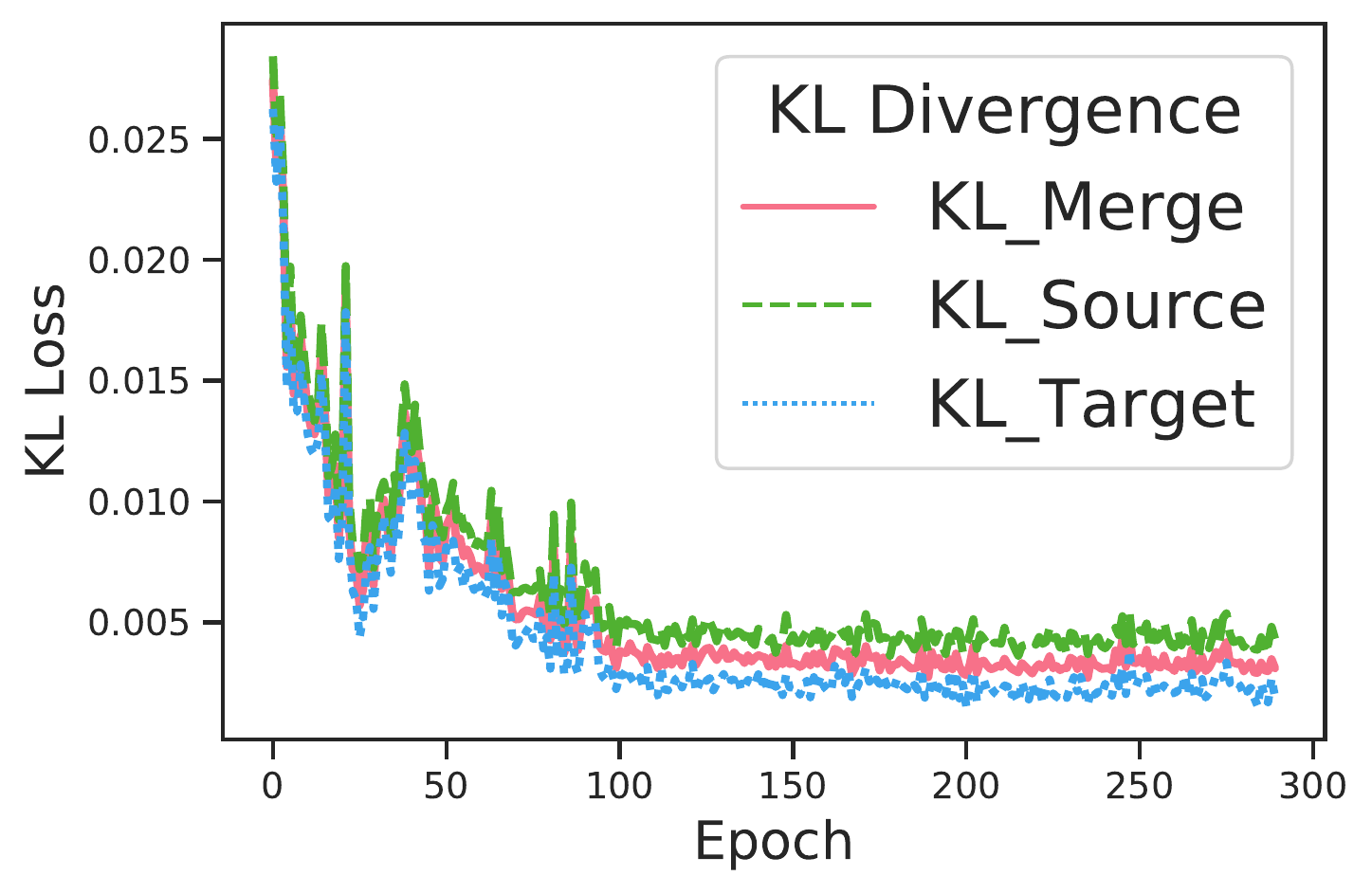}}
		\end{minipage}
		
		\caption{F1 Score and Loss on Twitter dataset under the guidance of KL divergence minimization.}
		\label{fig:loss_twitter}
	\end{figure}
	\section{Other Experiments}
	Due the space limitations, we only present the results of validation experiments on the Company dataset in the main body of this paper. And we also conduct comprehensive experiments on the Twitter dataset and present these results in the Appendix.
	\subsection{Supplementary Experiments for Sec. \ref{sec:exp_feat_completion}}
	The additional experiments to validate the effects of completion strategies on the Twitter dataset are shown in Table~\ref{tab:exp_completion_twitter}.
	\begin{table}[h]
		\centering
		\setlength{\tabcolsep}{3.0pt}
		\caption{Results of  models with different completion strategies on the Twitter dataset. "None" means we only use the observable  attributes (i.e., $X^o$) for both vocal and silent nodes without completion; "$0$-Completion" means we complete the missing features of silent nodes by zero vector; "Mean-of-Neighbors" strategy uses the mean vector of vocal neighbors to complete the missing dimensions for silent nodes.}
		\label{tab:exp_completion_twitter}
		\begin{tabular}{cccc}
			\toprule
			Dataset & Completion Method & \multicolumn{2}{c}{Twitter}\\
			\cmidrule(lr){3-4}
			Evaluation Metric & $\backslash$ & F1 &  AUC  \\
			\midrule
			\multirow{3} *{MLP} & None & 70.62\small{$\pm$1.01} & 78.11\small{$\pm$0.96} \\
			~& $\bm{0}$-Completion & 63.23\small{$\pm$0.8}5	& 69.7\small{$\pm$0.93}  \\
			~& Mean-of-Neighbors & 70.85\small{$\pm$1.31} &	80.12\small{$\pm$1.38} \\
			\hline
			\multirow{3} *{GCN} & None  & 79.03\small{$\pm$1.16} & 84.48\small{$\pm$0.57}	 \\
			~& $\bm{0}$-Completion & 78.03\small{$\pm$0.98} & 85.11\small{$\pm$1.81}	 \\
			~& Mean-of-Neighbors & 80.19\small{$\pm$0.87} & 86.88\small{$\pm$1.23}	 \\
			\hline
			\multirow{3} *{GraphSAGE} & None  & 83.99\small{$\pm$0.76 }& 92.19\small{$\pm$0.76}	\\
			~& $\bm{0}$-Completion & 85.40\small{$\pm$1.03} & 92.08\small{$\pm$1.61}	 \\
			~& Mean-of-Neighbors & 84.71\small{$\pm$0.92}  & 92.13\small{$\pm$1.53}	\\
			\hline
			\multirow{3} *{JKNet} & None  & 78.33\small{$\pm$1.96} &  85.53\small{$\pm$1.67} \\
			~& $\bm{0}$-Completion & 82.35\small{$\pm$1.31} & 89.52\small{$\pm$1.73}	\\
			~& Mean-of-Neighbors & 83.86\small{$\pm$1.03} & 91.17\small{$\pm$1.21} \\
			\hline
			\multirow{3} *{GCNII} & None  &77.65\small{$\pm$0.89} & 86.05\small{$\pm$1.13}	 \\
			~& $\bm{0}$-Completion & 80.23\small{$\pm$1.38} & 86.93\small{$\pm$1.51}	 \\
			~& Mean-of-Neighbors & 83.85\small{$\pm$1.17} &	90.67\small{$\pm$1.32} \\
			\hline 
			\multirow{3} *{OODGAT} & None  &75.65\small{$\pm$1.21} & 85.23\small{$\pm$2.21}	 \\
			~& $\bm{0}$-Completion & 80.58\small{$\pm$1.68} & 88.19\small{$\pm$1.52}	 \\
			~& Mean-of-Neighbors & 85.95\small{$\pm$2.01} &	92.67\small{$\pm$1.60} \\
			\midrule
			KTGNN & $\backslash$ & $\bm{89.65}$\small{$\bm{\pm1.20}$} & $\bm{95.08}$\small{$\bm{\pm0.93}$}  \\
			\bottomrule
		\end{tabular}
	\end{table}
	
	\subsection{Supplementary Experiments for Sec. \ref{sec:ablation}}
	The additional experiments to validate the effects of the KL loss term $\mathcal{L}^{kl}$ on the Twitter dataset are shown in Fig. \ref{fig:loss_twitter}. And the results on Silent-Graphs without knowledge transfer are shown in Table \ref{tab:silent_graph}.
	
        \begin{figure}[h]
		\begin{minipage}[t]{0.49\linewidth}
			\centering
			\subfloat[Twitter-$K$]{\includegraphics[width=\linewidth]{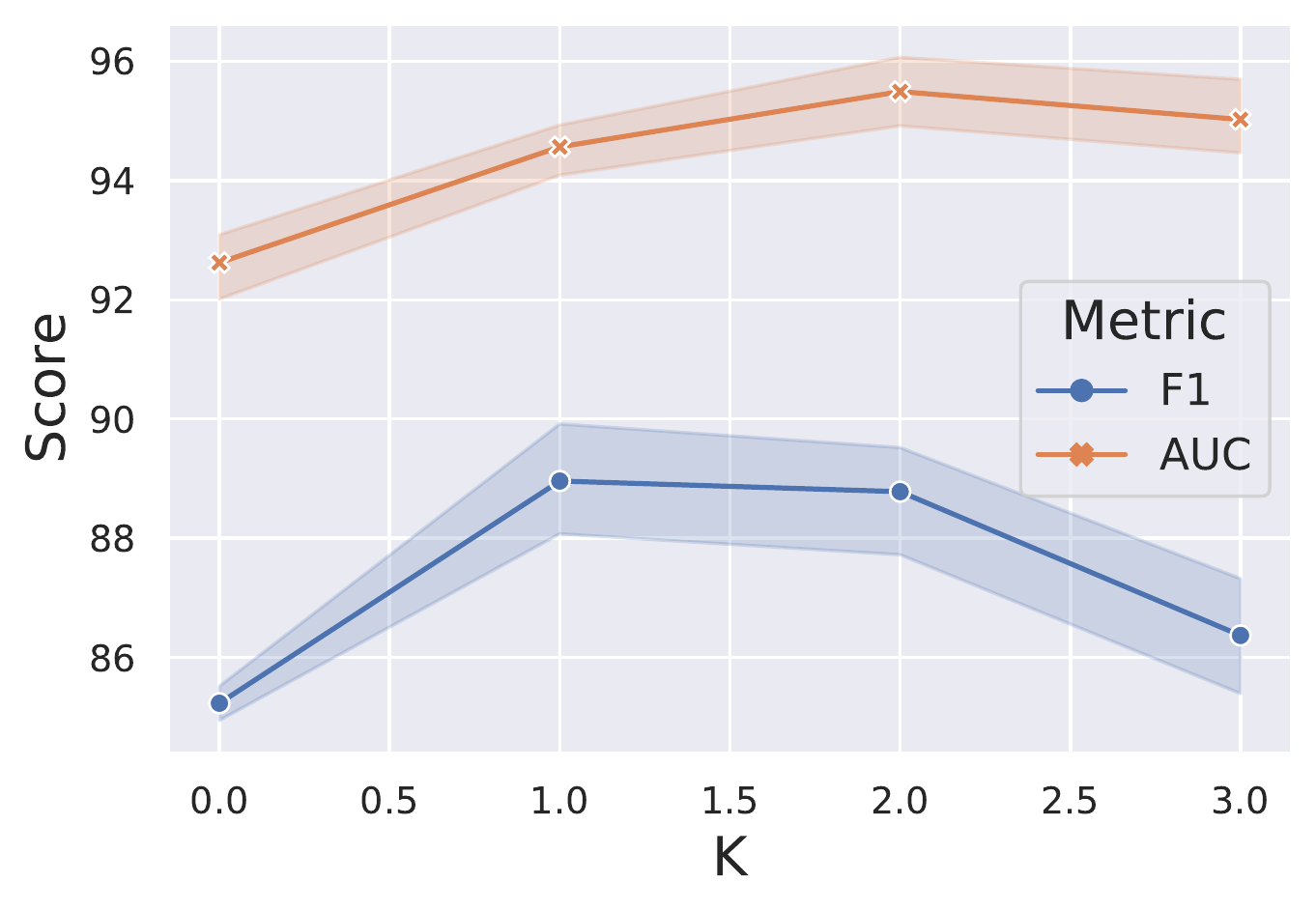}}
		\end{minipage}
		\begin{minipage}[t]{0.49\linewidth}
			\centering
			\subfloat[Twitter-$\lambda$]{\includegraphics[width=\linewidth]{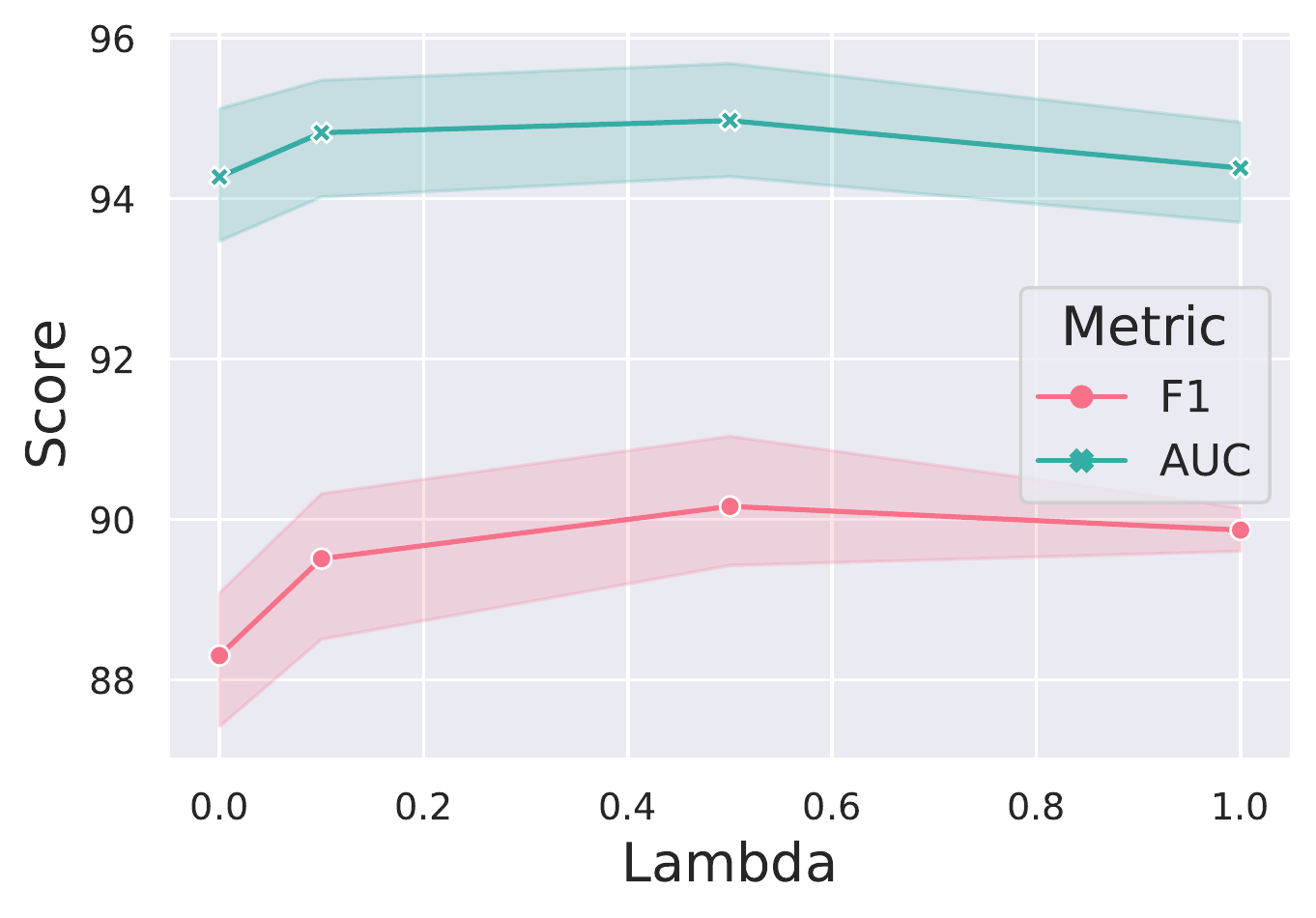}}
		\end{minipage}
		
		\caption{Hyper-parameter analysis on Twitter dataset.}
		\label{fig:hyper_twitter}
	\end{figure}
        \begin{table}[h]
		\centering
		\setlength{\tabcolsep}{3.0pt}
		\caption{Results on Silent-Graph (Company dataset), which removes all vocal nodes and their connected edges. Silent-Graph simulates the case without any knowledge transfer.}
		\label{tab:silent_graph}
		\begin{tabular}{cccc}
			\toprule
			Dataset & Input Graph & \multicolumn{2}{c}{Company}\\
			\cmidrule(lr){3-4}
			Evaluation Metric & $\backslash$ & F1 &  AUC  \\
			\midrule
			\multirow{2} *{GCN} & Silent-Graph & 55.70\small{$\pm$0.98} & 56.78\small{$\pm$0.88} \\
			~& VS-Graph & 57.60\small{$\pm0.49$} & 57.83\small{$\pm0.93$}  \\
			\hline
                \multirow{2} *{GraphSAGE} & Silent-Graph & 57.23\small{$\pm$0.83} & 57.71\small{$\pm$1.01} \\
			~& VS-Graph & 58.93\small{$\pm0.67$} & 60.63\small{$\pm0.57$}  \\
			\hline
                \multirow{2} *{JKNet} & Silent-Graph & 57.90\small{$\pm$1.32} & 58.11\small{$\pm$1.12} \\
			~& VS-Graph & 58.38\small{$\pm0.58$} & 61.06\small{$\pm0.53$}  \\
			\hline
                \multirow{2} *{GCNII} & Silent-Graph & 57.43\small{$\pm$1.21} & 58.39\small{$\pm$0.98} \\
			~& VS-Graph & 58.21\small{$\pm0.88$} & 60.06\small{$\pm0.63$}  \\
			\midrule
			KTGNN & VS-Graph & $\bm{64.96}$\small{$\bm{\pm0.63}$} & $\bm{67.11}$\small{$\bm{\pm0.52}$}  \\
			\bottomrule
		\end{tabular}
	\end{table}
 
	\subsection{Supplementary Experiments for Sec. \ref{sec:hyper_analysis}}
	The additional experiments to analyze the sensitivity of hyper-parameters on the Twitter dataset are shown in Fig.~\ref{fig:hyper_twitter}.
	
	\subsection{Supplementary Experiments for Sec. \ref{sec:exp_representation}}
	In the main body, we conduct T-SNE visualization on both the raw features and the representations learned by KTGNN. In the appendix, we also visualize the representations learned by vanilla GCN \cite{GCN} (initial missing features complemented by "Mean-of-Neighbors" strategy.), and the results show that GCN learns mixed representations that have lower distinguishability on nodes of different classes, and lose the domain difference between vocal nodes and silent nodes because GCN does not consider the distribution-shift between vocal nodes and silent nodes during message passing.
	
	\begin{figure}[h]
		\centering
		\includegraphics[width=\linewidth]{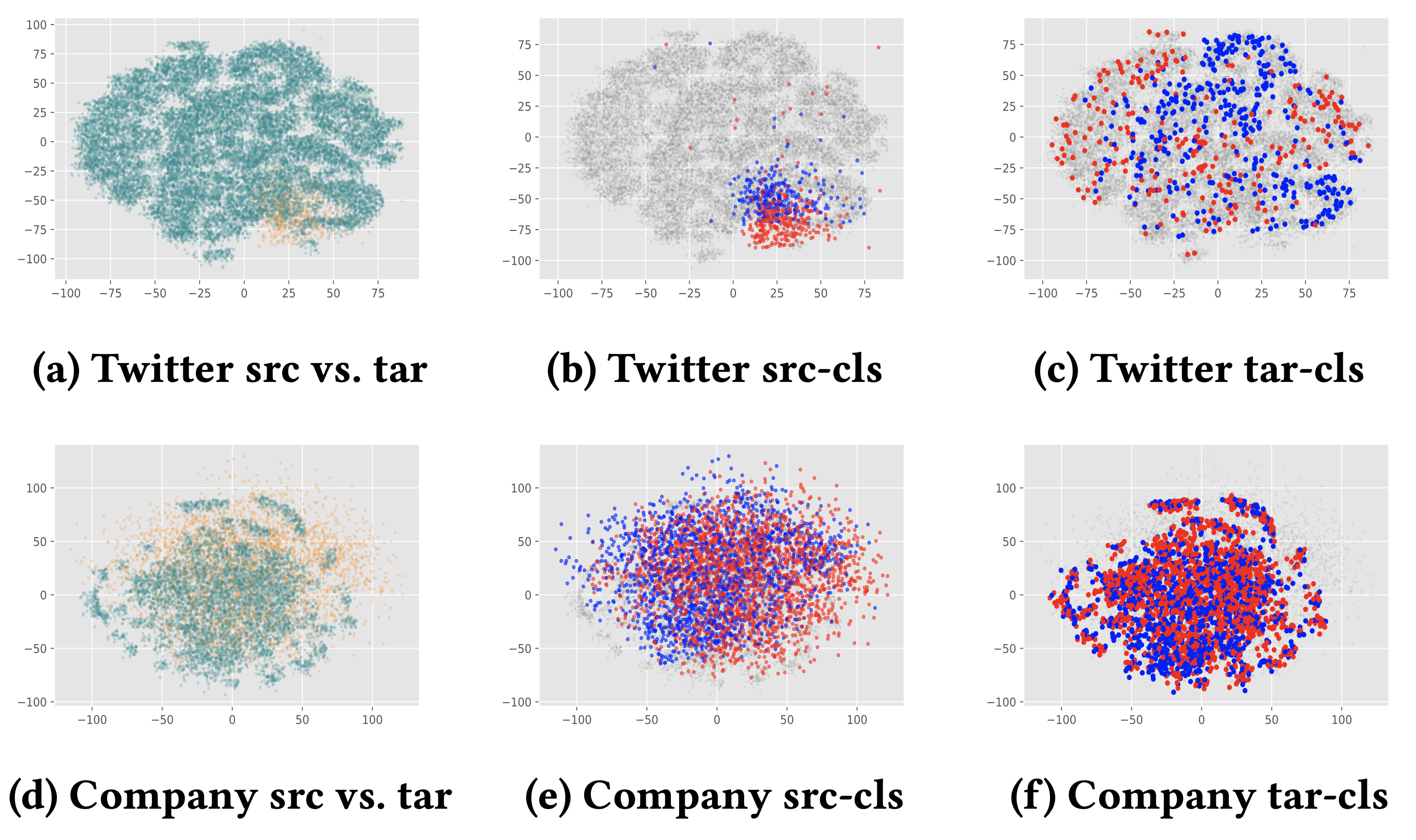}
		\caption{T-SNE visualization of the node completed representations learned by baseline GCN, distinguished by their populations (vocal$\rightarrow$orange, silent$\rightarrow$cyan) or node labels (binary classes: positive$\rightarrow$red, negative$\rightarrow$blue).}
		\label{fig:tsne_representation_gcn}
	\end{figure}
\section{Time Complexity Analysis of KTGNN}
We calculate the time complexity of KTGNN in this section. For simplification, we denote the dimension of the raw node feature and the hidden feature as $D$. We denote the set of edges among vocal nodes as $E^{v-v}$, the set of edges among silent nodes as $E^{s-s}$, and the set of edges between vocal nodes and silent nodes as $E^{v-s}$. Then the time complexity of the DAFC module is $\mathcal{O}\left((|E^{v-s}|+|E^{s-s}|)*D^2\right)$, the time complexity of the DAMP module is $\mathcal{O}\left(|E|*D^2\right)$ and the time complexity of the DTC module is $\mathcal{O}\left(|V|*D^2\right)$. Considering that $|E|=|E^{v-v}| + |E^{v-s}| + |E^{s-s}|$, \textbf{the final time complexity of KTGNN is $\mathcal{O}\left((|E|+|V|)*D^2\right)$, which equals to the complexity of other mainstream GNNs \cite{GAT, GATv2}}.

\section{Hyper-Parameter Settings}
    \label{appendix:hyper_parameter}
	For each dataset, we randomly divide the annotated silent nodes into train/valid/test sets with a fixed ratio of 60\%/20\%/20\%, and we add all annotated vocal nodes into the training set because our target is to classify the silent nodes. For KTGNN and each baseline model, we run 10 times with different random seeds.
	
	For fairness, we perform a hyper-parameter search for all models in the same searching space.  The hidden dimension of all models are searched in \{32, 64, 128, 256\} and we choose the number of training epochs from \{300, 600, 900\}. We use the Adam optimizer for all experiments and the learning rate is searched in \{1e-1, 5e-2, 1e-2, 5e-3, 1e-3, 5e-4, 1e-4\}, weight decay is searched in \{1e-4, 1e-3, 5e-3\}, and $\lambda$ (the weight of $\mathcal{L}^{kl}$) is searched in \{0.0, 0.1, 0.5, 1.0\}. For $\gamma$ (the weight of distribution consistency loss $\mathcal{L}^{dist}$ is fixed to 1.0, and fixed to 0.0 for the ablation study for $\text{KTGNN}_{\backslash \mathcal{L}^{dist}}$ which removes $\mathcal{L}^{dist}$ from KTGNN.  The number of layers for GNN models, including KTGNN and other baseline models except for GCNII and DAGNN, are searched in \{1, 2, 3, 4, 5, 6\}. The number of layers for GCNII and DAGNN (two deeper GCN models) are searched in \{2, 6, 10, 16, 20, 32, 64\} according to their papers \cite{GCN2, DAGNN}. For OODGAT, we treat the vocal nodes on the VS-Graph as its out-distribution nodes and treat the silent nodes as in-distribution nodes. And then OODGAT targets at detecting vocal nodes and then truncate the connections between vocal nodes and silent nodes. For other hyper-parameters of OODGAT, we use the recommended settings used in \citet{OODGAT}.  All models used in this paper were trained on Nvidia Tesla V100 (32G) GPU. 

\end{document}